\documentclass[10pt, twocolumn, letterpaper]{article}

\usepackage[pagenumbers]{iccv} 

\usepackage{bm} 
\usepackage{booktabs} 
\usepackage{colortbl} 
\usepackage{multirow, multicol, makecell} 
\usepackage{pifont} 
\usepackage[dvipsnames]{xcolor} 
\usepackage{chngcntr} 
\usepackage{nicematrix} 
\usepackage{caption} 
\usepackage{stfloats} 

\newcommand{\cmark}{\ding{51}}
\newcommand{\xmark}{\ding{55}}

\definecolor{tabfirst}{rgb}{1,0.7,0.7}
\definecolor{tabsecond}{rgb}{1,0.85,0.7}
\definecolor{tabthird}{rgb}{1,1,0.7}
\definecolor{darkred}{rgb}{0.6,0.16,0.25}
\definecolor{heavygray}{gray}{0.90}
\newcommand{\gain}[1]{\textcolor{red}{#1}}
\newcommand{\diff}[1]{\footnotesize\textcolor{red}{\textbf{(#1)}}}

\definecolor{iccvblue}{rgb}{0.21,0.49,0.74}
\usepackage[pagebackref, breaklinks, colorlinks, allcolors=iccvblue]{hyperref}
\usepackage[capitalize]{cleveref}

\def\paperID{3507} 
\def\confName{ICCV}
\def\confYear{2025}

\title{H3R: Hybrid Multi-view Correspondence for Generalizable 3D Reconstruction}

\author{ %
    Heng Jia$^{1,2}$ \quad Linchao Zhu$^{1,2}$ \quad Na Zhao$^{3}$ \\
    $^{1}$ReLER Lab, CCAI, Zhejiang University \quad
    $^{3}$Singapore University of Technology and Design \\
    $^{2}$The State Key Lab of Brain-Machine Intelligence, Zhejiang University
}

\begin{document}
\maketitle
\begin{abstract}
    Despite recent advances in feed-forward 3D Gaussian Splatting, generalizable 3D reconstruction remains challenging, particularly in multi-view correspondence modeling.
    Existing approaches face a fundamental trade-off: explicit methods achieve geometric precision but struggle with ambiguous regions, while implicit methods provide robustness but suffer from slow convergence.
    We present H3R, a hybrid framework that addresses this limitation by integrating volumetric latent fusion with attention-based feature aggregation.
    Our framework consists of two complementary components: an efficient latent volume that enforces geometric consistency through epipolar constraints, and a camera-aware Transformer that leverages Plücker coordinates for adaptive correspondence refinement.
    By integrating both paradigms, our approach enhances generalization while converging 2$\times$ faster than existing methods.
    Furthermore, we show that spatial-aligned foundation models (e.g., SD-VAE) substantially outperform semantic-aligned models (e.g., DINOv2), resolving the mismatch between semantic representations and spatial reconstruction requirements.
    Our method supports variable-number and high-resolution input views while demonstrating robust cross-dataset generalization.
    Extensive experiments show that our method achieves state-of-the-art performance across multiple benchmarks, with significant PSNR improvements of 0.59 dB, 1.06 dB, and 0.22 dB on the RealEstate10K, ACID, and DTU datasets, respectively.
    Code is available at \url{https://github.com/JiaHeng-DLUT/H3R}.
\end{abstract}
\section{Introduction}
\label{sec:intro}

\begin{figure}[!htbp]
    \centering
    \includegraphics[width=0.8\linewidth]{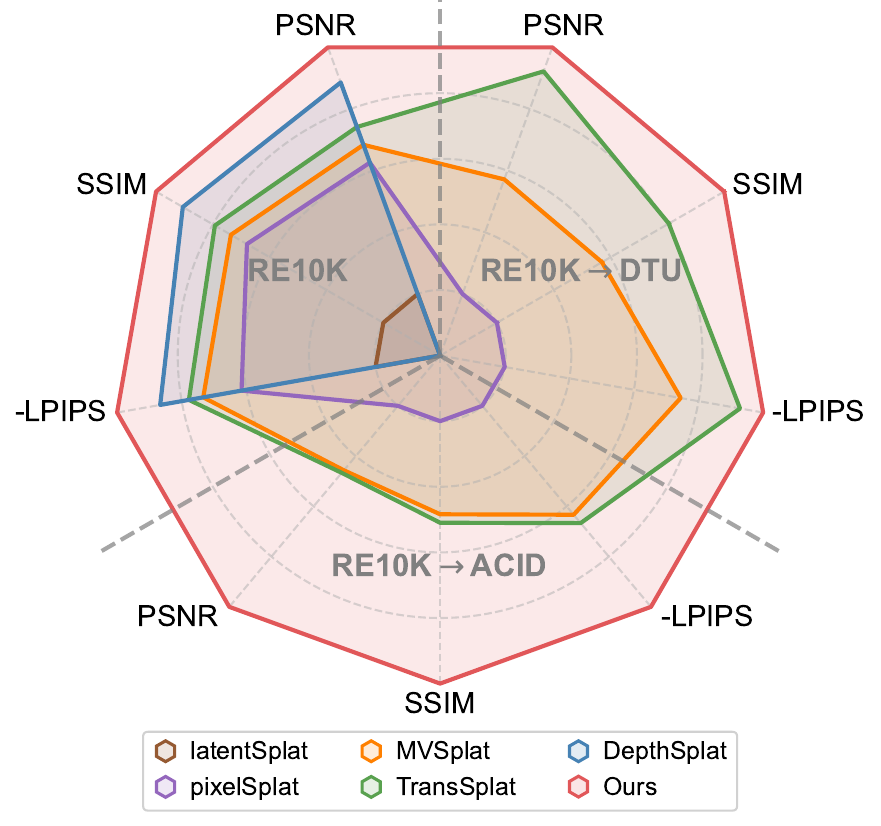}
    \caption{
        \textbf{Comparison with state-of-the-art generalizable 3D reconstruction methods.}
        Our method consistently outperforms existing approaches across three datasets, demonstrating superior reconstruction performance and cross-dataset generalization.
    }
    \label{fig:radar}
    \vspace{-1em}
\end{figure}

Generalizable 3D reconstruction~\cite{charatan2024pixelsplat,chen2024mvsplat,tang2024hisplat,zhang2024transplat,min2024epipolar,zhang2024gs,chen2024mvsplat360,wang2021ibrnet,wang2021neus,liu2024mvsgaussian,chan2023generative,xie2024lrm,hong2023lrm,trevithick2021grf,chen2021mvsnerf,liu2022neural,sajjadi2022scene,johari2022geonerf,cong2023enhancing,liu2023semantic,jiang2023leap,li2024gp,wang2024gs2} has attracted significant attention for its ability to recover 3D scenes without per-scene optimization.
Traditional Neural Radiance Fields (NeRF) require costly scene-specific training, limiting its practicality.
Recent advances~\cite{charatan2024pixelsplat,chen2024mvsplat,tang2024hisplat,zhang2024transplat,min2024epipolar,zhang2024gs,chen2024mvsplat360,liu2024mvsgaussian} shift toward feed-forward reconstruction methods that leverage efficient representations like 3D Gaussian Splatting (3DGS)~\cite{kerbl20233d,wang2024view}.
3DGS represents scenes as collections of 3D Gaussian primitives, bypassing the computationally intensive ray-marching of NeRF and enabling fast rasterization-based rendering.
pixelSplat~\cite{charatan2024pixelsplat} and MVSplat~\cite{chen2024mvsplat} predict pixel-aligned Gaussian primitives in a single forward pass, achieving high-quality reconstructions on real-world benchmarks.
Concurrently, Transformer-based approaches like eFreeSplat~\cite{min2024epipolar} and TranSplat~\cite{zhang2024transplat} improve cross-view correspondence through attention mechanisms, enhancing depth estimation from sparse inputs.
Despite substantial progress in reconstruction quality, significant challenges remain in achieving robust generalizable 3D reconstruction across diverse scenes.

The \textbf{first challenge} lies in \textit{the trade-off between explicit and implicit correspondence modeling.
}
Explicit methods~\cite{chen2024mvsplat,wang2024freesplat,liu2024mvsgaussian,fei2024pixelgaussian,tang2024hisplat,zheng2024gps} enforce geometric constraints through cost volumes, achieving high geometric precision but struggling when photometric consistency breaks down, \textit{e.g.}, under occlusions, textureless regions, specular highlights, or repetitive patterns~\cite{zhang2024transplat}.
In contrast, implicit methods~\cite{liu2023epipolar,johari2022geonerf,suhail2022generalizable,du2023learning,jin2024lvsm,hong2023lrm,xu2024grm,zhang2024gs} employ attention mechanisms to learn robust correspondences for ambiguous scenarios, but suffer from slow convergence~\cite{zhang2024gs,xu2024grm,tang2025lgm,hong2023lrm}.
Explicit methods deliver geometric precision but lack robustness, while implicit methods provide adaptability at the cost of efficiency.
Therefore, a principled integration of these paradigms is critical for robust and efficient multi-view correspondence modeling.

We present \textbf{H3R}, \textit{a hybrid network that integrates explicit and implicit correspondence modeling}.
It comprises two complementary components: volumetric latent fusion that enforces geometric consistency through epipolar constraints, and camera-aware Transformer that performs guided correspondence aggregation across viewpoints.
The volumetric module projects multi-view features into a discretized latent volume that explicitly captures geometric correspondences.
The Transformer leverages Plücker coordinates~\cite{royal1864philosophical} to perform geometry-informed attention, implicitly modeling cross-view correspondences.
This hybrid design achieves superior reconstruction quality while significantly improving training efficiency.

Beyond architectural design, we find that the choice of visual representation plays a critical role in determining correspondence quality.
This leads us to identify a \textbf{second key challenge}: \textit{the mismatch between the semantic nature of prevailing visual representations and the spatial fidelity required for 3D reconstruction}.
Recent methods~\cite{tang2024hisplat,charatan2024pixelsplat,chen2025lara,kulhanek2024wildgaussians,tang2025lgm,shen2025gamba,zhang2024geolrm,jena2025sparsplat,cao2024mvsformer++} typically rely on semantic-aligned models such as DINO~\cite{oquab2023dinov2,caron2021emerging}, which emphasize high-level semantic understanding through global image-level supervision.
However, this comes at the cost of pixel-level spatial fidelity, which is crucial for accurate 3D reconstruction.
In contrast, spatial-aligned models (such as MAE~\cite{he2022masked} and SD-VAE~\cite{rombach2022high}) preserve fine-grained local features and geometric structures through pixel-level reconstruction objectives, making them inherently more suitable for 3D reconstruction tasks that require precise cross-view correspondence.
Through systematic evaluation across diverse visual foundation models (\cref{fig:encoder_performance}), we demonstrate that \textit{spatial-aligned models consistently outperform semantic-aligned counterparts}.
Notably, SD-VAE achieves the optimal performance with superior parameter efficiency, while DINOv2 substantially underperforms despite widespread adoption.
These results provide strong evidence that spatial-aligned models are superior for generalizable 3D reconstruction.

Beyond these technical considerations, practical deployment presents challenges.
The \textbf{third challenge} stems from \textit{the variability of real-world inputs.
}
Existing methods assume controlled settings with fixed views and uniform quality, struggling with diverse real-world scenarios.
To address this deployment gap, we develop \textit{two specialized extensions} of our base H3R model.
H3R-$\alpha$ processes multiple context views through adaptive latent fusion, consistently improving reconstruction quality with additional views (\cref{fig:view_scaling}).
H3R-$\beta$ specializes in high-resolution reconstruction, delivering high-fidelity results with limited input views (\cref{tab:re10k_acid}).

Our key contributions are summarized as follows:
\begin{itemize}
    \item \textbf{Hybrid Multi-view Correspondence:}
          We introduce H3R, a hybrid framework that resolves the trade-off between explicit and implicit correspondence modeling.
          By unifying explicit volumetric fusion and implicit camera-aware Transformer, H3R achieves superior reconstruction quality across challenging scenes and converges 2$\times$ faster than state-of-the-art methods.
    \item \textbf{Spatial-aligned Visual Representation:}
          We systematically demonstrate that spatial-aligned visual representations significantly outperform conventional semantic-aligned features for high-fidelity 3D reconstruction.
          Notably, SD-VAE achieves the optimal performance with superior parameter efficiency, while DINOv2 substantially underperforms despite widespread adoption.
    \item \textbf{Comprehensive Performance Improvements:}
          We address three core challenges in generalizable 3D reconstruction, achieving state-of-the-art reconstruction quality (\cref{fig:radar}), strong generalization (\cref{tab:cross_dataset}), 2$\times$ faster convergence (\cref{tab:ablation}) and robust real-world adaptability (\cref{fig:view_scaling}).
          We demonstrate significant PSNR improvements of 0.59 dB, 1.06 dB, and 0.22 dB on RealEstate10K, ACID, and DTU datasets respectively.
\end{itemize}

\section{Related work}

\noindent\textbf{Generalizable 3D Gaussian Splatting.}
Recent 3DGS-based methods substantially improve the quality of generalizable 3D reconstruction.
Object-centric methods~\cite{szymanowicz2024splatter,chen2025lara,jianggaussianblock,xu2024grm,tang2025lgm,zhang2024gs,zou2024triplane}, trained on large-scale datasets~\cite{deitke2023objaverse,deitke2023objaversexl,qiu2024richdreamer}, achieve robust object reconstruction from sparse views and demonstrate strong generalization across diverse object categories.
In parallel, scene-level methods~\cite{charatan2024pixelsplat,chen2024mvsplat,zhang2024gs,zhang2024transplat,min2024epipolar,zheng2024gps} predict pixel-aligned Gaussian parameters to reconstruct unbounded scenes.
For example, pixelSplat~\cite{charatan2024pixelsplat} integrates epipolar Transformer with probabilistic depth sampling to resolve scale ambiguities.
GPS-Gaussian~\cite{zheng2024gps} and MVSplat~\cite{chen2024mvsplat} leverage cost volumes to improve geometric accuracy.
eFreeSplat~\cite{min2024epipolar} enforces depth consistency through iterative Gaussian alignment, and TranSplat~\cite{zhang2024transplat} refines Gaussian centers using monocular depth priors and depth-aware deformable matching.
Despite these advances, existing methods are often limited to sparse input views (1-4) and struggle in challenging real-world scenarios.
Therefore, we introduce a hybrid framework that accepts multi-view inputs, produces high-quality reconstructions, and exhibits strong generalization.

\noindent\textbf{Multi-view Stereo.}
Multi-view stereo (MVS) reconstructs 3D structure from a collection of calibrated 2D images~\cite{collins1996space} by exploiting either \textit{explicit} or \textit{implicit} multi-view correspondences.
Explicit cost volume methods~\cite{yao2018mvsnet,gu2020casmvsnet,ding2022transmvsnet,peng2022unimvsnet,xu2023unifying,xiong2023cl} build plane-sweep stereo volumes~\cite{collins1996space,yao2018mvsnet} to aggregate multi-view correspondences along epipolar lines, thus achieving efficient reconstruction.
However, they struggle in challenging scenarios where photometric consistency assumptions break down, such as occlusions, textureless surfaces, specular reflections, and repetitive patterns.
Conversely, attention-based methods~\cite{liu2023epipolar,johari2022geonerf,suhail2022generalizable,du2023learning,jin2024lvsm,hong2023lrm,xu2024grm,zhang2024gs} learn implicit correspondences via attention mechanisms, but often suffer from slow convergence~\cite{zhang2024gs,xu2024grm,tang2025lgm,hong2023lrm}.
To address these limitations, we propose a hybrid framework that combines explicit and implicit correspondences, leading to fast convergence and robust reconstruction across diverse scenes.

\noindent\textbf{Visual Foundation Models.}
Visual foundation models can be categorized by their supervision paradigm into \textit{semantic-aligned} and \textit{spatial-aligned} models.
Semantic-aligned models focus on global semantic understanding through image-level supervision.
While effective at capturing high-level holistic representations, these models often sacrifice spatial fidelity and visual details.
Representative examples include DeiT III~\cite{touvron2022deit} for image classification and CLIP~\cite{radford2021learning} for vision-language alignment.
In contrast, spatial-aligned models learn explicit spatial structures and fine-grained visual cues through pixel-aligned supervision.
Examples include Depth Anything~\cite{yang2024depth,yang2024depthv2} for depth estimation, SAM~\cite{kirillov2023segment,ravi2024sam} for semantic segmentation, DUSt3R~\cite{wang2024dust3r} and MASt3R~\cite{leroy2025grounding} for 3D point regression and matching, as well as MAE~\cite{he2022masked} and SD-VAE~\cite{rombach2022high} for image reconstruction.
Our experiments demonstrate that spatial-aligned backbones consistently outperform semantic-aligned counterparts, underscoring the importance of spatial-aligned supervision for generalizable 3D reconstruction.

\section{Method}

\begin{figure*}[!htbp]
    \centering
    \includegraphics[width=1.0\textwidth]{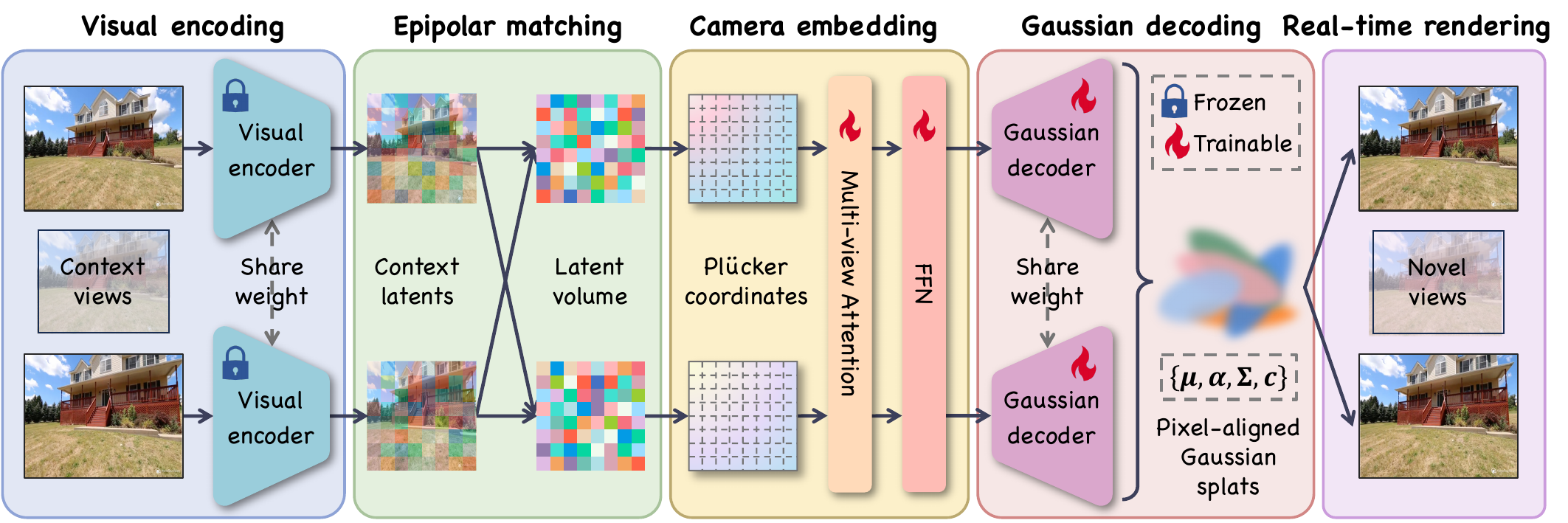}
    \caption{
        \textbf{Overview of our generalizable 3D reconstruction framework H3R.}
        Our method generates pixel-aligned Gaussians without per-scene optimization, achieving high-fidelity 3D reconstruction.
        The framework comprises five key stages: \textbf{(i) Visual encoding}: We employ a visual foundation model to extract rich latents from each context view, encoding appearance and structural information essential for 3D reconstruction.
        \textbf{(ii) Epipolar matching}: These latents are then aligned and aggregated into a coherent latent volume using epipolar geometric constraints, establishing spatial correspondences across views.
        \textbf{(iii) Camera embedding}: To further strengthen these correspondences, we incorporate camera parameters via Plücker coordinates and apply multi-view attention to refine the latents.
        \textbf{(iv) Gaussian decoding}: A lightweight CNN-based decoder subsequently transforms these refined latents into pixel-aligned Gaussians.
        \textbf{(v) Real-time rendering}: Finally, the generated Gaussian splats enable high-fidelity, real-time novel view synthesis.
    }
    \label{fig:framework}
    \vspace{-1em}
\end{figure*}

We propose H3R (\cref{fig:framework}), a hybrid multi-view correspondence framework for generalizable 3D reconstruction.
H3R combines volumetric latent fusion with a camera-aware Transformer, achieving superior rendering quality and accelerated convergence.
To handle varying inputs in practical 3D reconstruction scenarios, we develop two extensions of our base model.
H3R-$\alpha$ handles multiple context views and integrates target view poses to ensure comprehensive scene coverage, while H3R-$\beta$ specializes in high-resolution reconstruction and delivers high-quality results even with few input views.
We present the base H3R model in~\cref{sec:sparse_view_reconstruction} and detail the multi-view extension H3R-$\alpha$ in~\cref{sec:multiview_reconstruction}.

\subsection{Sparse-view Reconstruction}
\label{sec:sparse_view_reconstruction}

Given $N$ sparse context images $\{I_{i} \in \mathbb{R}^{H \times W \times 3}\}_{i=1}^{N}$, we first employ a visual foundation model $\mathcal{E}$ to independently encode each image $I_{i}$:
\begin{equation}
    \bm{x}_{i} = \mathcal{E}(I_{i}) \in \mathbb{R}^{h \times w \times c}.
\end{equation}
The core challenge lies in effectively fusing these per-view features $\{\bm{x}_i\}_{i=1}^N$ to generate a unified 3D Gaussian representation of the scene:
\begin{equation}
    f \colon \{\bm{x}_{i}\}_{i=1}^{N} \to \{\bm{G}_{j}\}_{j=1}^{N \times H \times W \times K},
\end{equation}
where $K$ represents the number of Gaussians per pixel.
We introduce two complementary modules designed to model cross-view relationships: volumetric latent fusion and camera-aware Transformer.

\noindent\textbf{Volumetric Latent Fusion.}
We construct latent volumes using epipolar geometry to capture cross-view correspondences.
Following plane-sweep stereo~\cite{collins1996space,yao2018mvsnet,im2019dpsnet,xu2023unifying}, we sample depth planes and warp neighboring view latents into the reference coordinate system.
Within the near and far depth bounds, we uniformly sample $d$ depth planes $\{\bm{D}^{k}\}_{k=1}^{d}$ in inverse depth space.
For each reference view $i$, we warp the latent $\bm{x}_{j}$ from neighboring view $j$ onto these depth planes via differentiable homography~\cite{yao2018mvsnet}:
\begin{equation}
    \bm{x}_{j \to i}^{k} = \text{Warp}(\bm{x}_{j}, \mathbf{P}_{i \to j}, \bm{D}^{k}) \in \mathbb{R}^{h \times w \times c},
\end{equation}
where $\mathbf{P}_{i \to j}$ denotes the relative camera projection from view $i$ to $j$.
We then concatenate the warped latents from all depth planes to construct the latent volume $\bm{x}_{j \to i}$:
\begin{equation}
    \bm{x}_{j \to i} = \left[\bm{x}_{j \to i}^{1}, \bm{x}_{j \to i}^{2}, \dots, \bm{x}_{j \to i}^{d}\right] \in \mathbb{R}^{h \times w \times d \times c},
\end{equation}
where $\left[\cdot\right]$ denotes concatenation along the depth dimension.

We evaluate three strategies for latent volume construction: \textit{correlation-based}, \textit{difference-based}, and \textit{cost-free}.
The \textit{correlation-based} approach, widely adopted in multi-view stereo~\cite{yao2018mvsnet}, measures feature similarity using the scaled dot product between reference features $\bm{x}_{i}$ and warped source features $\bm{x}_{j \to i}$:
\begin{equation}
    \bm{x}_{j \to i}^{\prime} = \frac{\bm{x}_{i} \cdot \bm{x}_{j \to i}}{\sqrt{c}} \in \mathbb{R}^{h \times w \times d},
\end{equation}
where $\bm{x}_{j \to i}^{\prime}$ represents the constructed latent volume.
This formulation produces distinct peaks at correct correspondences, particularly in texture-rich regions.
The \textit{difference-based} approach constructs the volume by concatenating element-wise feature differences across all $d$ depth planes:
\begin{equation}
    \bm{x}_{j \to i}^{\prime} = \bm{x}_{j \to i} - \bm{x}_{i} \in \mathbb{R}^{h \times w \times (d \times c)}.
\end{equation}
This strategy excels at localizing correspondences along sharp edges where feature disparities are most salient.
The cost-free approach forgoes explicit similarity metrics, directly concatenating warped features across depth planes:
\begin{equation}
    \bm{x}_{j \to i}^{\prime} = \bm{x}_{j \to i} \in \mathbb{R}^{h \times w \times (d \times c)}.
\end{equation}
This design preserves the complete feature representations, allowing subsequent layers to learn task-specific matching functions.
Our experiments (\cref{tab:cost_function}) demonstrate that the cost-free strategy achieves superior cross-dataset generalization compared to handcrafted methods.

To integrate multi-view information, we fuse original latent $\bm{x}_{i}$ with latent volume $\bm{x}_{i}^{\prime}$ through linear projections:
\begin{equation}
    \bm{z}_{i} = \mathrm{Linear}_{1}\left(\bm{x}_{i}\right) + \mathrm{Linear}_{2}\left(\bm{x}_{j \to i}^{\prime}\right) \in \mathbb{R}^{h \times w \times c^{\prime}},
\end{equation}
where $\mathrm{Linear}_{1}\colon \mathbb{R}^{c}\to \mathbb{R}^{c^{\prime}}$ and $\mathrm{Linear}_{2}\colon \mathbb{R}^{d \times c}\to \mathbb{R}^{c^{\prime}}$ project the latents into a shared $c^{\prime}$-dimensional space.

\noindent\textbf{Camera-aware Transformer.}
Camera parameters provide essential geometric constraints for 3D scene understanding, reconstruction, and reasoning~\cite{xu2023dmv3d,hong2023lrm,shi2023mvdream,liu2023zero}.
Pixel-aligned camera rays encode rich geometric priors that vary across spatial locations and viewpoints.
We leverage this information using Plücker ray coordinates~\cite{royal1864philosophical,xu2023dmv3d,zhang2024gs,tang2025lgm,xu2024grm}:
\begin{equation}
    \bm{r}_{i}=[\boldsymbol{o}_{i} \otimes \boldsymbol{d}_{i}, \boldsymbol{d}_{i}] \in \mathbb{R}^{h \times w \times 6},
\end{equation}
where $\boldsymbol{o}_{i}$ and $\boldsymbol{d}_{i}$ denote the origin and direction of each pixel ray for view $i$, and $\otimes$ represents cross product.

The visual tokens $\{\bm{z}_{i}\}_{i=1}^{N}$ and Plücker coordinates $\{\bm{r}_{i}\}_{i=1}^{N}$ from all input views are flattened and concatenated into 1D sequences $\bm{z} \in \mathbb{R}^{(N \times h \times w) \times c^{\prime}}$ and $\bm{r} \in \mathbb{R}^{(N \times h \times w) \times 6}$.
The Plücker coordinates are embedded and fused with visual tokens to enable camera conditioning.
Self-attention layers implicitly strengthen multi-view correlations, followed by a feed-forward network (FFN).
The process is formalized as:
\begin{equation}
    \begin{aligned}
        \bm{z} & = \bm{z} + \text{PosEmb}(\bm{r}),                      \\
        \bm{z} & = \bm{z} + \text{Attention}(\text{LayerNorm}(\bm{z})), \\
        \bm{z} & = \bm{z} + \text{FFN}(\text{LayerNorm}(\bm{z})).
    \end{aligned}
\end{equation}
Here, $\text{PosEmb}\colon \mathbb{R}^{6}\to \mathbb{R}^{c^{\prime}}$ maps Plücker coordinates to the feature dimension.

\noindent
\textbf{Gaussian Decoding.}
After processing through the camera-aware Transformer, we upsample the fused multi-view latents to the original input resolution and decode them into Gaussian parameters using a hierarchical CNN decoder $\mathcal{D}$:
\begin{equation}
    \bm{G} = \mathcal{D}(\bm{z}) \in \mathbb{R}^{N \times H \times W \times K}.
\end{equation}
The decoder $\mathcal{D}$ employs sequential upsampling ResBlocks~\cite{he2016deep}.
Following~\cite{zhang2024gs}, Gaussian primitives are parameterized by 3-channel RGB, 3-channel scale, 4-channel rotation quaternion, 1-channel opacity, and 1-channel ray distance.
For rendering, Gaussian center positions are derived from ray distances and camera parameters.
See Appendix~\cref{sec:gaussian_parameterization} for more details.

\noindent
\textbf{Training Loss.}
We employ mean-squared error ($\text{MSE}$) and learned perceptual image patch similarity ($\text{LPIPS}$)~\cite{zhang2018unreasonable} losses to supervise novel view synthesis, following~\cite{charatan2024pixelsplat,tang2024hisplat,chen2024mvsplat,zhang2024transplat,hong2023lrm,tang2025lgm,xu2024grm,zhang2024gs}.
We also incorporate mean-absolute-error ($\text{MAE}$) loss on image gradients to enhance photometric consistency.
The overall training objective is defined as:
\begin{equation}
    \label{eq:loss}
    \mathcal{L}_{\mathcal{D}} = \text{MSE}(I, \hat{I}) + \lambda \cdot \text{LPIPS}(I, \hat{I}) + \text{MAE}(\nabla I, \nabla \hat{I}),
\end{equation}
where $I$ and $\hat{I}$ denote the ground truth and rendered images, respectively, $\nabla$ is the gradient operator, and $\lambda = 0.05$ balances the $\text{LPIPS}$ loss.

\subsection{Multi-view Reconstruction}
\label{sec:multiview_reconstruction}

\noindent
\textbf{Handling Variable Input Views.}
To accommodate varying numbers of input views in real-world 3D reconstruction scenarios, we train our model on diverse view conditions and incorporate target view camera poses into the Gaussian generation process.
For multiple context views, we compute pixel-wise average of the warped latents during volumetric fusion, enabling adaptive input handling:
\begin{equation}
    \bm{x}_{i}^{\prime} = \frac{1}{N-1}\sum_{j \neq i}\bm{x}_{j\to i}^{\prime},
\end{equation}
where $\bm{x}_{i}^{\prime}$ denotes the fused latent volume for view $i$.
Our camera-aware Transformer architecture inherently supports dynamic input view counts.
As shown in~\cref{fig:view_scaling}, reconstruction quality consistently improves with increased views, demonstrating adaptability across varying input views.

\noindent
\textbf{Target View Integration.}
Existing generalizable 3DGS methods do not explicitly generate Gaussians for target views, limiting comprehensive scene reconstruction.
We address this limitation by incorporating target view camera poses (\textit{without corresponding images}) to generate pixel-aligned Gaussians for target views.
We use zero tensors for target view visual features while computing valid Plücker coordinates from their camera poses, feeding both into the Transformer.
This facilitates information exchange between context and target poses, enhancing Gaussian generation for complete view coverage.
A challenge is that without an explicit supervision, the model tends to ignore the target pose information during training.
To address this, we introduce an auxiliary reconstruction head that predicts target images from final-layer features, supervised by~\cref{eq:loss}.
This ensures target-view Gaussians actively participate in rendering, achieving comprehensive scene coverage.
Qualitative results are shown in Appendix~\cref{fig:target_pose}.

\section{Experiments}

\subsection{Main Results}

\noindent\textbf{Setup.}
H3R is pre-trained using two context views at a 256$\times$256 resolution.
We subsequently fine-tune two variants from this base model: 1) H3R-$\alpha$ supports 2-8 input views at the 256$\times$256 resolution and 2) H3R-$\beta$ accepts two context views at a higher 512$\times$512 resolution.
H3R-$\alpha$ also accepts target camera poses as input to generate Gaussians with comprehensive view coverage, which substantially improves rendering quality in unobserved regions.
See Appendix~\cref{sec:implementation_details} for more details.

\begin{table}[!htbp]
    \centering
    \resizebox{\linewidth}{!}{
        \setlength{\tabcolsep}{2pt}
        \begin{tabular}{l|c|ccc|ccc}
            \toprule                                     %
            \multirow{2}{*}{\textbf{Method}}            & \textbf{\#Gaussians}               & \multicolumn{3}{c|}{\textbf{RealEstate10K}~\cite{zhou2018stereo}} & \multicolumn{3}{c}{\textbf{ACID}~\cite{liu2021infinite}}                                                                                                                             \\
                                                        & ($N \times H \times W \times K$)   & \textbf{PSNR$\uparrow$}                                           & \textbf{SSIM$\uparrow$}                                  & \textbf{LPIPS$\downarrow$}   & \textbf{PSNR$\uparrow$}      & \textbf{SSIM$\uparrow$}      & \textbf{LPIPS$\downarrow$}   \\
            \midrule                                     %
            pixelNeRF~\cite{yu2021pixelnerf}            & \multirow{4}{*}{N/A}               & 20.43                                                             & 0.589                                                    & 0.550                        & 20.97                        & 0.547                        & 0.533                        \\
            GPNR~\cite{suhail2022generalizable}         &                                    & 24.11                                                             & 0.793                                                    & 0.255                        & 25.28                        & 0.764                        & 0.332                        \\
            AttnRend~\cite{du2023learning}              &                                    & 24.78                                                             & 0.820                                                    & 0.213                        & 26.88                        & 0.799                        & 0.218                        \\
            MuRF~\cite{xu2024murf}                      &                                    & 26.10                                                             & 0.858                                                    & 0.143                        & 28.09                        & 0.841                        & 0.155                        \\
            \midrule                                     %
            latentSplat~\cite{wewer2025latentsplat}     & $2 \times 256 \times 256 \times 3$ & 23.88                                                             & 0.812                                                    & 0.164                        & -                            & -                            & -                            \\
            pixelSplat~\cite{charatan2024pixelsplat}    & $2 \times 256 \times 256 \times 3$ & 26.09                                                             & 0.863                                                    & 0.136                        & 28.27                        & 0.843                        & 0.146                        \\
            MVSplat~\cite{chen2024mvsplat}              & $2 \times 256 \times 256 \times 1$ & 26.39                                                             & 0.869                                                    & 0.128                        & 28.25                        & 0.843                        & 0.144                        \\
            MVSplat360~\cite{chen2024mvsplat360}        & $2 \times 256 \times 256 \times 1$ & 26.41                                                             & 0.869                                                    & 0.126                        & -                            & -                            & -                            \\
            eFreeSplat~\cite{min2024epipolar}           & $2 \times 256 \times 256 \times 1$ & 26.45                                                             & 0.865                                                    & 0.126                        & 28.30                        & 0.851                        & \cellcolor{tabfirst}{0.140}  \\
            TranSplat~\cite{zhang2024transplat}         & $2 \times 256 \times 256 \times 1$ & 26.69                                                             & 0.875                                                    & 0.125                        & \cellcolor{tabthird}{28.35}  & 0.845                        & 0.143                        \\
            DepthSplat$^{\dag}$~\cite{xu2024depthsplat} & -                                  & 27.44                                                             & \cellcolor{tabthird}{0.887}                              & 0.119                        & -                            & -                            & -                            \\
            \midrule                                     %
            H3R                                         & $2 \times 256 \times 256 \times 1$ & \cellcolor{tabthird}{27.60}                                       & \cellcolor{tabsecond}{0.891}                             & \cellcolor{tabthird}{0.114}  & 28.29                        & \cellcolor{tabthird}{0.846}  & \cellcolor{tabthird}{0.144}  \\
            H3R-$\alpha$                                & $5 \times 256 \times 256 \times 1$ & \cellcolor{tabsecond}{27.79}                                      & \cellcolor{tabsecond}{0.891}                             & \cellcolor{tabsecond}{0.113} & \cellcolor{tabsecond}{28.44} & \cellcolor{tabsecond}{0.847} & \cellcolor{tabthird}{0.144}  \\
            H3R-$\beta$                                 & $2 \times 512 \times 512 \times 1$ & \cellcolor{tabfirst}{28.03}                                       & \cellcolor{tabfirst}{0.897}                              & \cellcolor{tabfirst}{0.110}  & \cellcolor{tabfirst}{28.71}  & \cellcolor{tabfirst}{0.853}  & \cellcolor{tabsecond}{0.141} \\
            \bottomrule
        \end{tabular}
    }
    \caption{
        \textbf{Quantitative evaluation.}
        Each model takes two context images with camera parameters as input and synthesizes three novel views for assessment.
        H3R-$\alpha$ processes multiple context images at 256$\times$256 alongside target-view cameras, while H3R-$\beta$ is optimized for two high-resolution context images at 512$\times$512.
        We highlight the \colorbox{tabfirst}{best}, \colorbox{tabsecond}{second-best}, and \colorbox{tabthird}{third-best} results.
        $^{\dag}$DepthSplat~\cite{xu2024depthsplat} uses additional pre-training data.
    }
    \label{tab:re10k_acid}
\end{table}

\begin{figure*}[!htbp]
    \centering
    \begin{subfigure}[t]{0.49\linewidth}
        \centering
        \includegraphics[width=\linewidth]{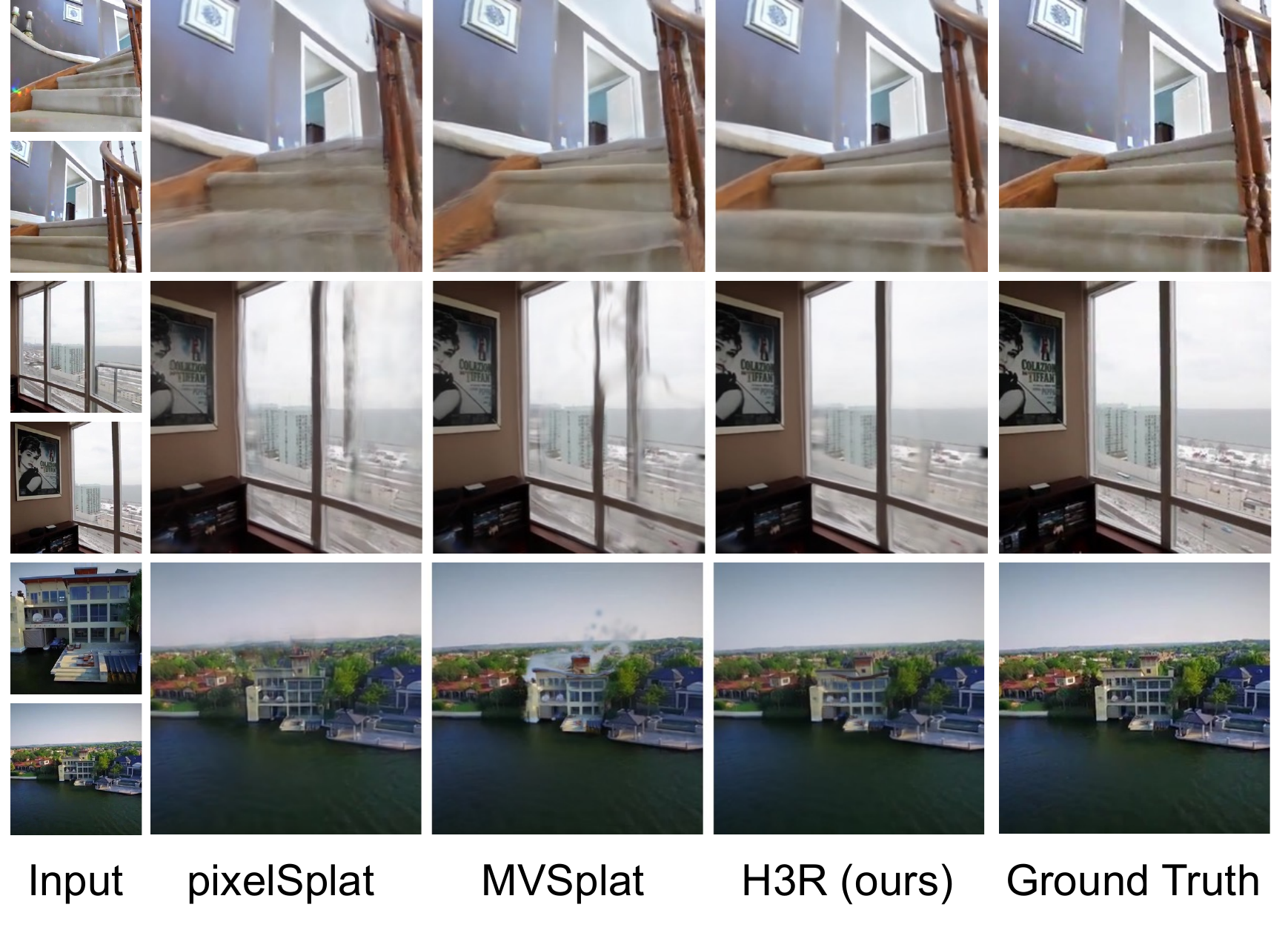}
    \end{subfigure}
    \begin{subfigure}[t]{0.49\linewidth}
        \centering
        \includegraphics[width=\linewidth]{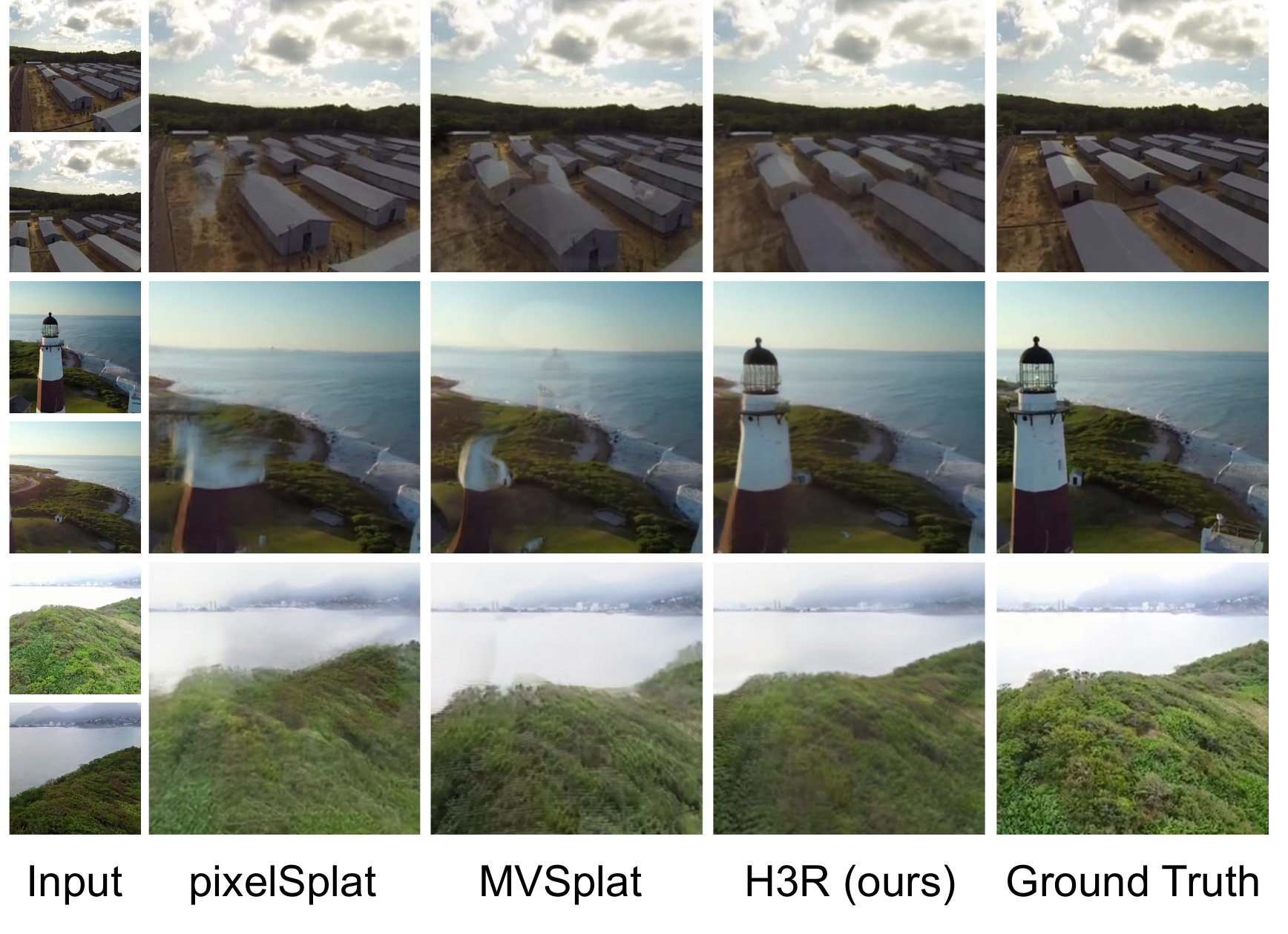}
    \end{subfigure}
    \caption{
        \textbf{Qualitative results on RealEstate10K (left) and ACID (right).} \textbf{See Appendix~\cref{fig:target_pose,fig:view_scaling_visual,fig:high_resolution} for more visualizations.}
    }
    \label{fig:re10k_acid}
\end{figure*}

\noindent\textbf{Two-view Novel View Synthesis.}
We evaluate our approach against state-of-the-art methods on two challenging datasets: RealEstate10K~\cite{zhou2018stereo} and ACID~\cite{liu2021infinite}, as presented in~\cref{tab:re10k_acid}.
On RealEstate10K, H3R outperforms TranSplat~\cite{zhang2024transplat} by a significant margin \gain{(+0.91 PSNR, +0.016 SSIM, and -0.011 LPIPS)}.
On ACID, our method achieves a PSNR of 28.29 and SSIM of 0.846, surpassing both pixelSplat~\cite{charatan2024pixelsplat} and MVSplat~\cite{chen2024mvsplat}.
The quantitative results, along with qualitative comparisons in~\cref{fig:re10k_acid}, demonstrate our method's effectiveness in producing high-fidelity novel views from sparse inputs.

\begin{figure*}[!htbp]
    \centering
    \begin{subfigure}
        {0.33\textwidth}
        \centering
        \includegraphics[width=\linewidth]{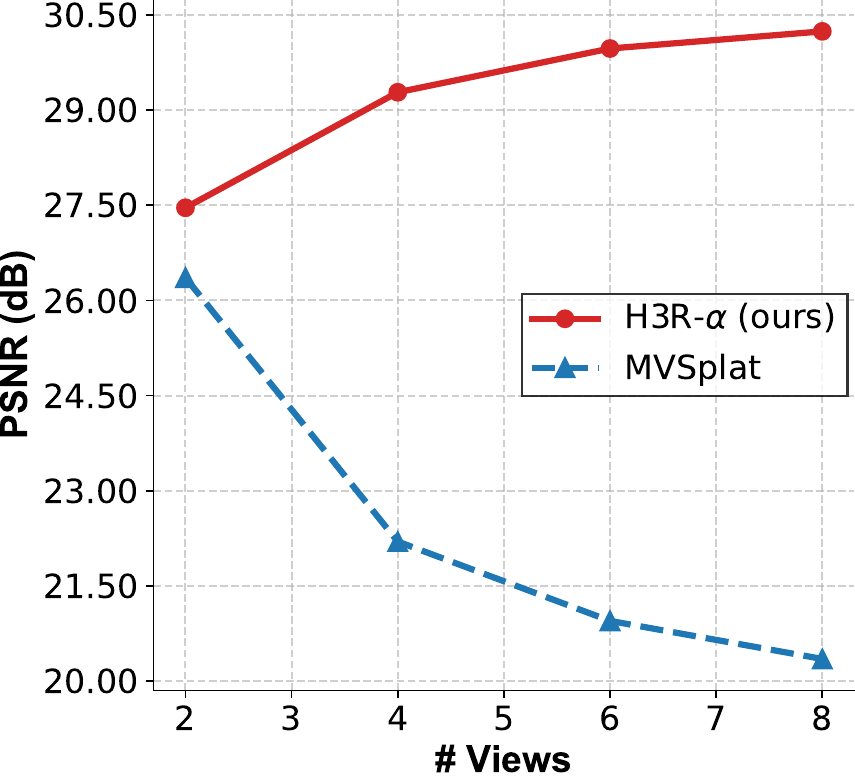}
        \label{fig:view_scaling_psnr}
    \end{subfigure}
    \begin{subfigure}
        {0.33\textwidth}
        \centering
        \includegraphics[width=\linewidth]{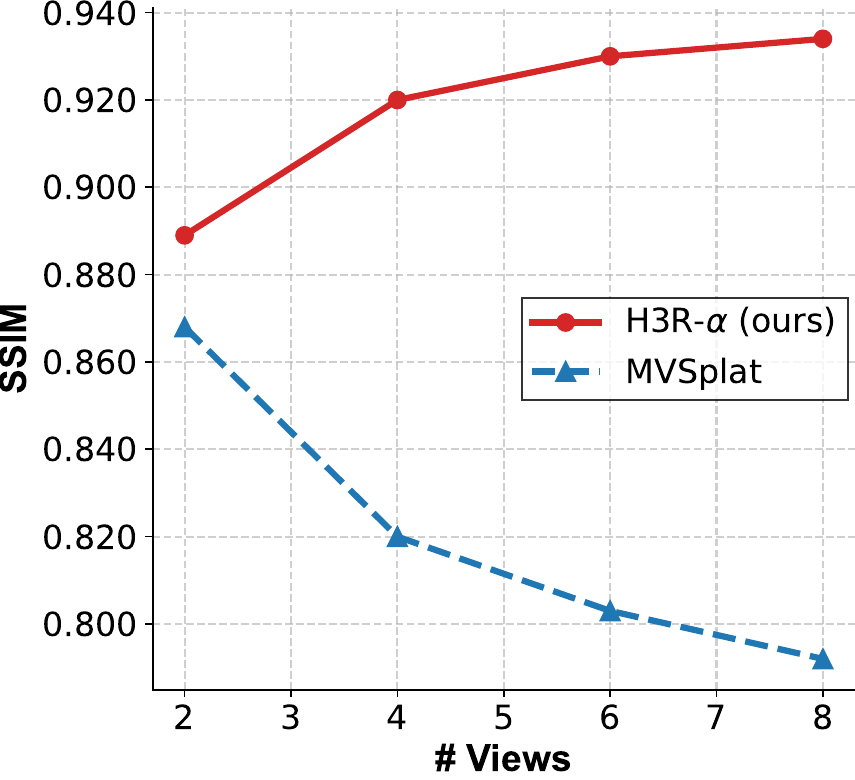}
        \label{fig:view_scaling_ssim}
    \end{subfigure}
    \begin{subfigure}
        {0.33\textwidth}
        \centering
        \includegraphics[width=\linewidth]{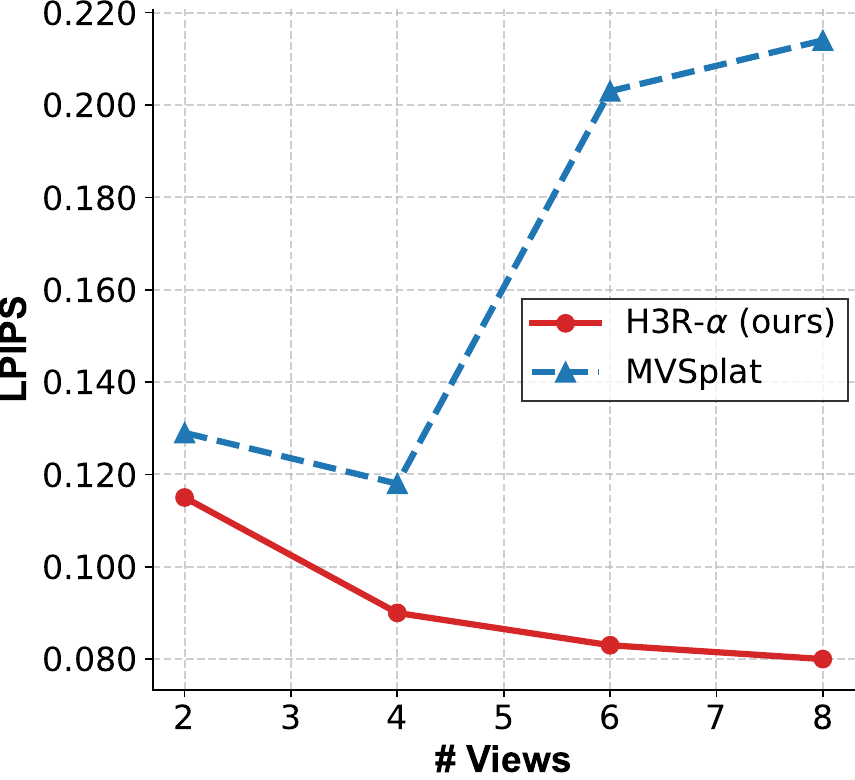}
        \label{fig:view_scaling_lpips}
    \end{subfigure}
    \vspace{-2em}
    \caption{
        \textbf{Comparison with MVSplat~\cite{chen2024mvsplat} across varying input
            views on RealEstate10K.}
        Our method outperforms MVSplat across all metrics and scales better with additional input views, while MVSplat's performance declines.
        See Appendix~\cref{tab:view_scaling} for detailed results.
    }
    \label{fig:view_scaling}
\end{figure*}

Our approach exhibits versatility through two specialized variants, each tailored for distinct input conditions.
H3R-$\alpha$ leverages target camera poses to generate Gaussians with comprehensive view coverage, achieving consistent improvements over the base H3R model (\gain{+0.19 and +0.15 PSNR} on RealEstate10K and ACID, respectively).
H3R-$\beta$ exploits higher-resolution inputs for enhanced detail capture, achieving superior rendering quality (\gain{+0.43 and +0.42 PSNR}).
These results demonstrate our approach's versatility across diverse input conditions, advancing state-of-the-art in generalizable novel view synthesis and highlighting potential for real-world deployment.

\noindent\textbf{Multi-view Novel View Synthesis.}
\cref{fig:view_scaling} compares the performance scaling trends of H3R-$\alpha$ with MVSplat~\cite{chen2024mvsplat} across varying number of input views on RealEstate10K.
H3R-$\alpha$ consistently improves with additional views, achieving \gain{+2.78 PSNR and -0.035 LPIPS} gains when scaling from 2 to 8 views.
In contrast, MVSplat~\cite{chen2024mvsplat} shows performance degradation (\gain{-6.01 PSNR}) in perceptual quality (\gain{+0.085 LPIPS}).
At 8 views, the performance gap reaches \gain{+9.89 PSNR and -0.134 LPIPS}.
The monotonic improvement across all metrics demonstrates H3R-$\alpha$'s superior adaptability for practical multi-view reconstruction scenarios.

\begin{figure*}[!tbp]
    \centering
    \begin{subfigure}[t]{0.49\linewidth}
        \centering
        \includegraphics[width=\linewidth]{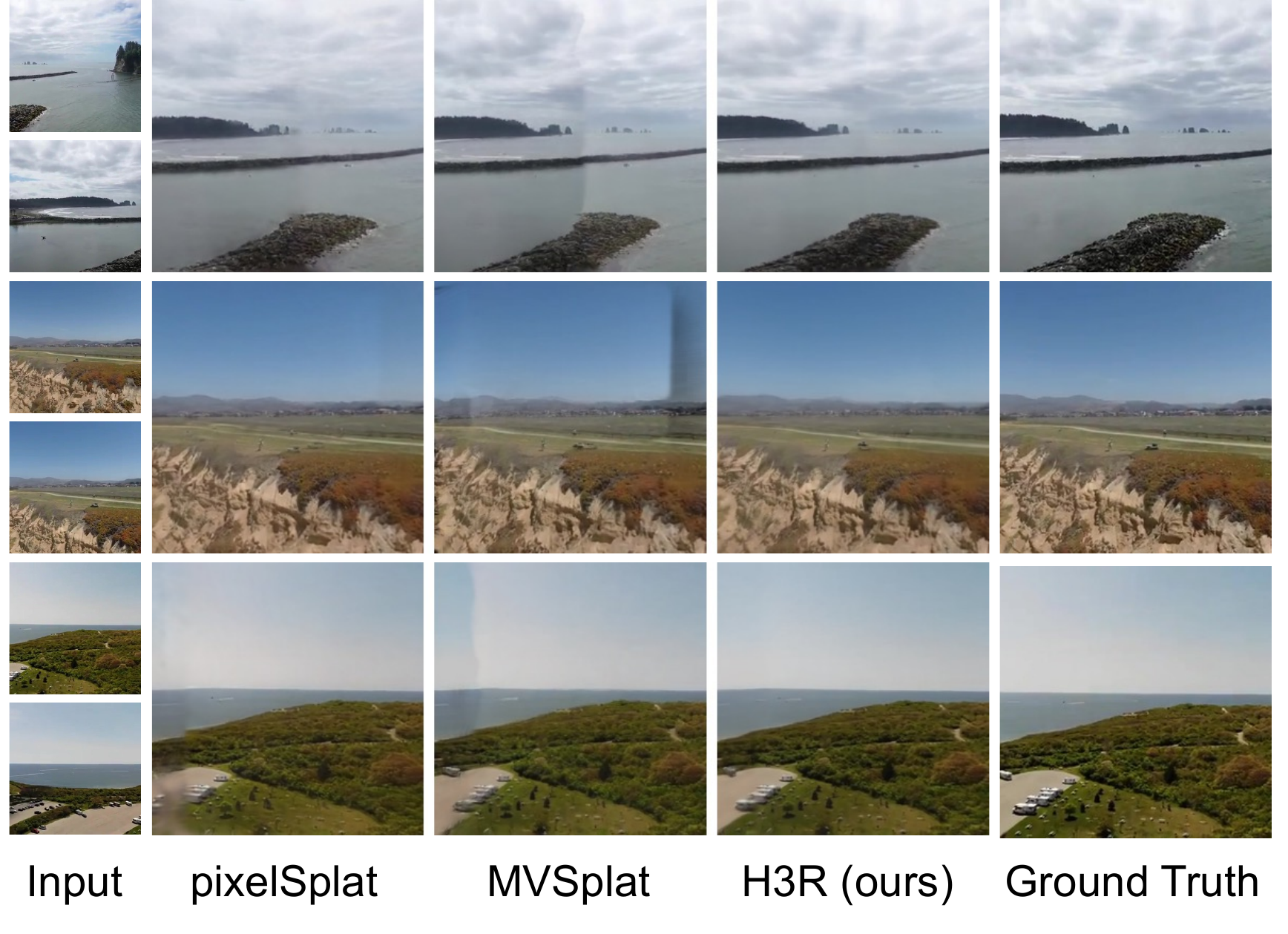}
    \end{subfigure}
    \begin{subfigure}[t]{0.49\linewidth}
        \centering
        \includegraphics[width=\linewidth]{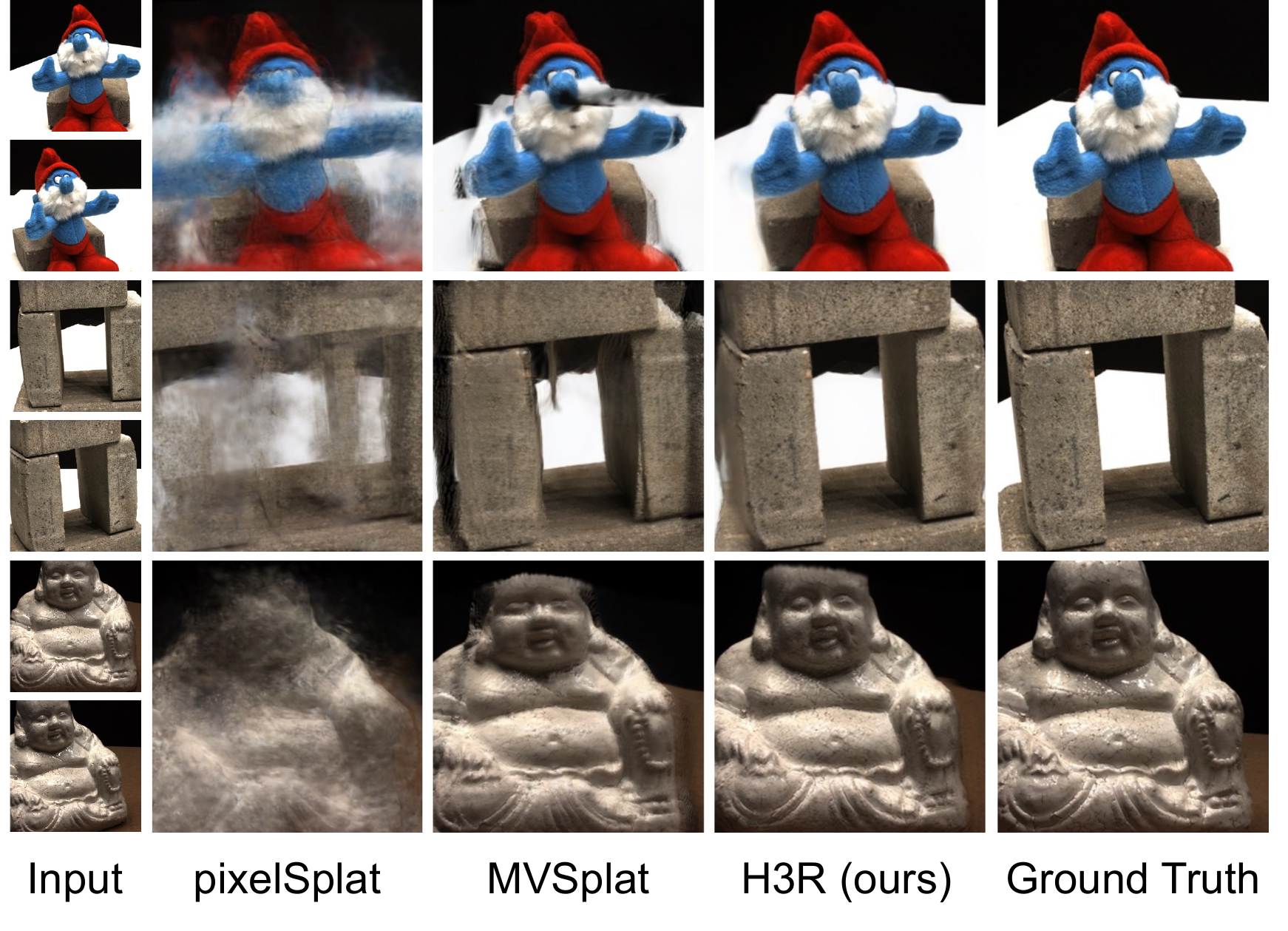}
    \end{subfigure}
    \caption{
        \textbf{Cross-dataset generalization from indoor RealEstate10K to outdoor ACID (left) and object-centric DTU (right).}
    }
    \label{fig:re10k2acid_dtu}
\end{figure*}

\noindent\textbf{Cross-dataset Generalization.}
We evaluate cross-dataset generalization by training on RealEstate10K (indoor) and performing zero-shot evaluation on ACID (outdoor) and DTU (object-centric).
As shown in~\cref{tab:cross_dataset}, our approach consistently outperforms existing methods across both datasets.
Compared to TranSplat~\cite{zhang2024transplat}, H3R achieves substantial improvements of \gain{+0.62 PSNR, +0.012 SSIM, and -0.004 LPIPS} on ACID, and \gain{+0.22 PSNR, +0.062 SSIM, and -0.021 LPIPS} on DTU.
~\cref{fig:re10k2acid_dtu} demonstrates that our method produces high-fidelity outputs while baselines suffer from artifacts.
Our model variants H3R-$\alpha$ and H3R-$\beta$ further enhance performance, demonstrating robust generalization across diverse scene types.

\begin{figure*}[!htbp]
    \centering
    \begin{subfigure}
        {0.33\textwidth}
        \centering
        \includegraphics[width=\linewidth]{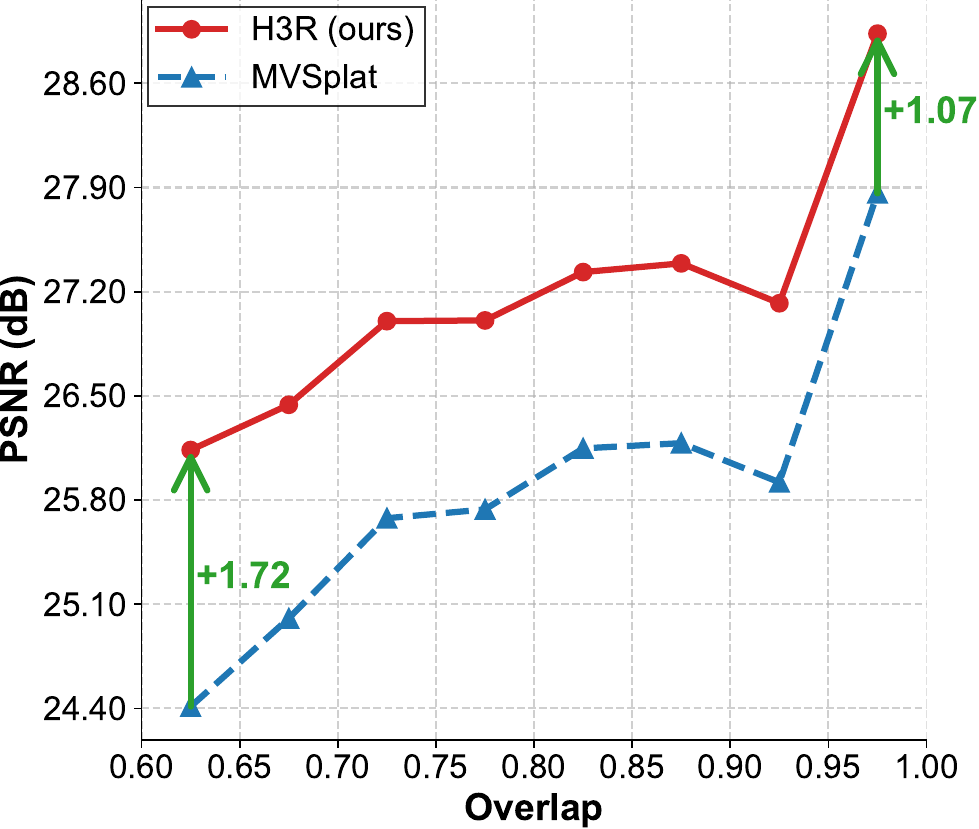}
    \end{subfigure}
    \begin{subfigure}
        {0.33\textwidth}
        \centering
        \includegraphics[width=\linewidth]{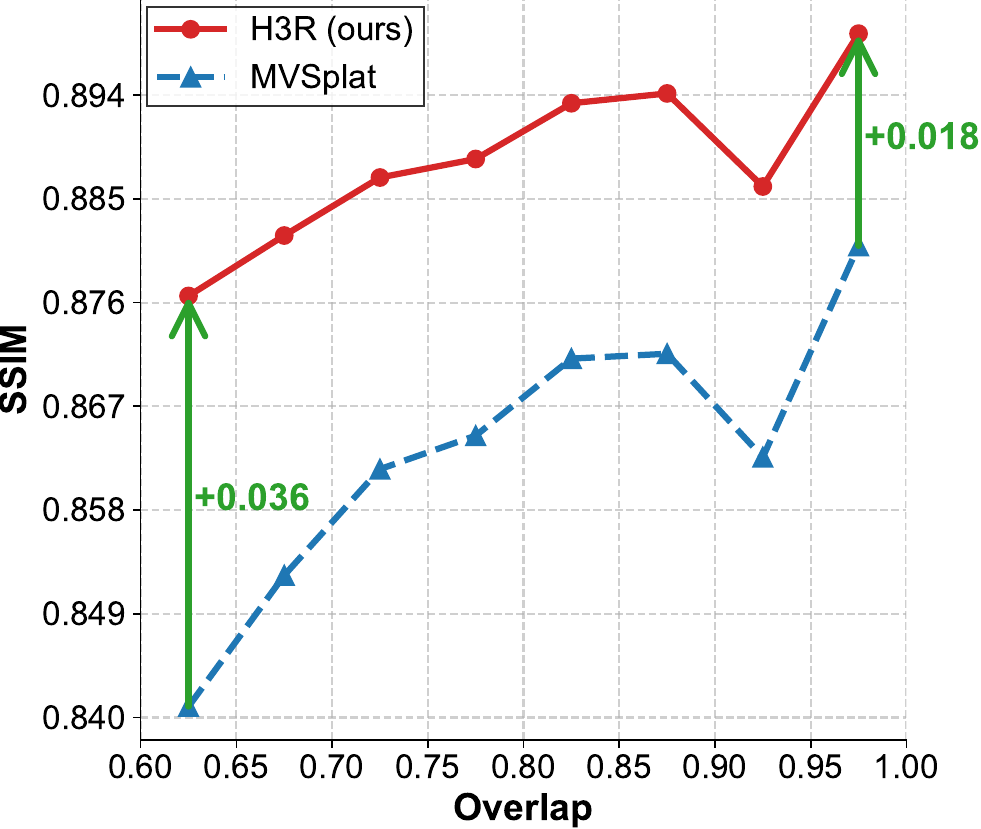}
    \end{subfigure}
    \begin{subfigure}
        {0.33\textwidth}
        \centering
        \includegraphics[width=\linewidth]{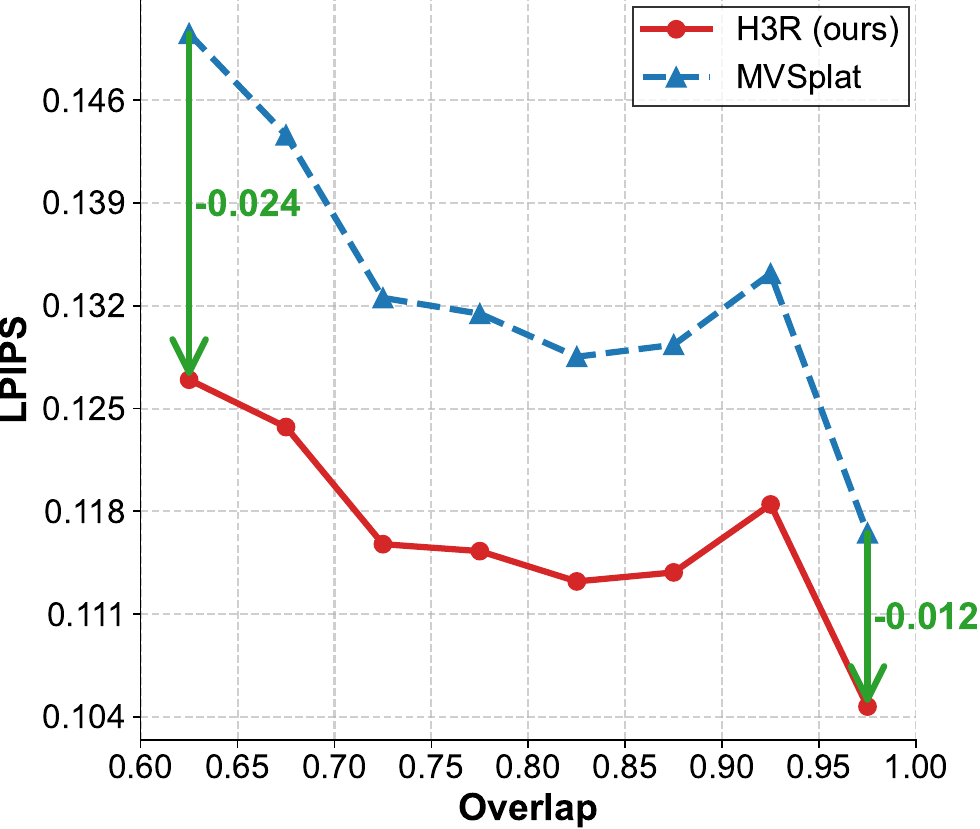}
    \end{subfigure}
    \caption{
        \textbf{Comparison with MVSplat~\cite{chen2024mvsplat} across varying overlaps on RealEstate10K.}
        Our method consistently outperforms MVSplat across all overlaps, with particularly significant improvements under low-overlap scenarios.
        See Appendix~\cref{tab:overlap} for detailed results.
    }
    \label{fig:overlap}
\end{figure*}

\noindent\textbf{Robustness to varying overlaps.}
We partition the RealEstate10K dataset into overlap-based subsets to evaluate our model's robustness.
As illustrated in~\cref{fig:overlap}, our approach consistently outperforms MVSplat~\cite{chen2024mvsplat} across all overlap ranges, especially under low-overlap scenarios.
More specifically, our method achieves substantial gains of \gain{+1.72 PSNR, +0.036 SSIM, and -0.024 LPIPS} under the lowest overlap condition.
The robust performance stems from our integration of explicit volumetric latent fusion with implicit camera-aware Transformer, which effectively establishes strong cross-view relationships even when epipolar priors become insufficient.

\begin{table}[!tbp]
    \centering
    \resizebox{\linewidth}{!}{
        \setlength{\tabcolsep}{2pt}
        \begin{tabular}{l|ccc|ccc|ccc}
            \toprule
            \multirow{2}{*}{\textbf{Method}}         & \multicolumn{3}{c|}{\textbf{ACID}} & \multicolumn{3}{c|}{\textbf{DTU (2 context views)}} & \multicolumn{3}{c}{\textbf{DTU (3 context views)}}                                                                                                                                                                                           \\
                                                     & \textbf{PSNR$\uparrow$}            & \textbf{SSIM$\uparrow$}                             & \textbf{LPIPS$\downarrow$}                         & \textbf{PSNR$\uparrow$}      & \textbf{SSIM$\uparrow$}      & \textbf{LPIPS$\downarrow$}   & \textbf{PSNR$\uparrow$}      & \textbf{SSIM$\uparrow$}      & \textbf{LPIPS$\downarrow$}   \\
            \midrule
            pixelSplat~\cite{charatan2024pixelsplat} & 27.64                              & 0.830                                               & 0.160                                              & 12.89                        & 0.382                        & 0.560                        & 12.52                        & 0.367                        & 0.585                        \\
            MVSplat~\cite{chen2024mvsplat}           & 28.15                              & 0.841                                               & 0.147                                              & 13.94                        & 0.473                        & 0.385                        & 14.30                        & 0.508                        & 0.371                        \\
            TransSplat~\cite{zhang2024transplat}     & 28.17                              & 0.842                                               & 0.146                                              & \cellcolor{tabthird}{14.93}  & 0.531                        & 0.326                        & -                            & -                            & -                            \\
            \midrule
            H3R                                      & \cellcolor{tabthird}{28.79}        & \cellcolor{tabsecond}{0.854}                        & \cellcolor{tabthird}{0.142}                        & \cellcolor{tabsecond}{15.15} & \cellcolor{tabsecond}{0.593} & \cellcolor{tabsecond}{0.305} & \cellcolor{tabthird}{15.27}  & \cellcolor{tabsecond}{0.600} & \cellcolor{tabthird}{0.323}  \\
            H3R-$\alpha$                             & \cellcolor{tabsecond}{28.91}       & \cellcolor{tabsecond}{0.854}                        & \cellcolor{tabsecond}{0.141}                       & \cellcolor{tabfirst}{15.17}  & \cellcolor{tabfirst}{0.605}  & \cellcolor{tabthird}{0.309}  & \cellcolor{tabfirst}{15.98}  & \cellcolor{tabfirst}{0.634}  & \cellcolor{tabfirst}{0.302}  \\
            H3R-$\beta$                              & \cellcolor{tabfirst}{29.23}        & \cellcolor{tabfirst}{0.861}                         & \cellcolor{tabfirst}{0.136}                        & \cellcolor{tabsecond}{15.15} & \cellcolor{tabthird}{0.579}  & \cellcolor{tabfirst}{0.303}  & \cellcolor{tabsecond}{15.36} & \cellcolor{tabthird}{0.599}  & \cellcolor{tabsecond}{0.315} \\
            \bottomrule
        \end{tabular}
    }
    \caption{
        \textbf{Cross-dataset generalization.}
        Models trained on RealEstate10K (indoor) demonstrate strong zero-shot performance on ACID (outdoor) and DTU (object).
    }
    \label{tab:cross_dataset}
    \vspace{-1em}
\end{table}
\begin{figure*}[!htbp]
    \centering
    \includegraphics[width=1.0\textwidth]{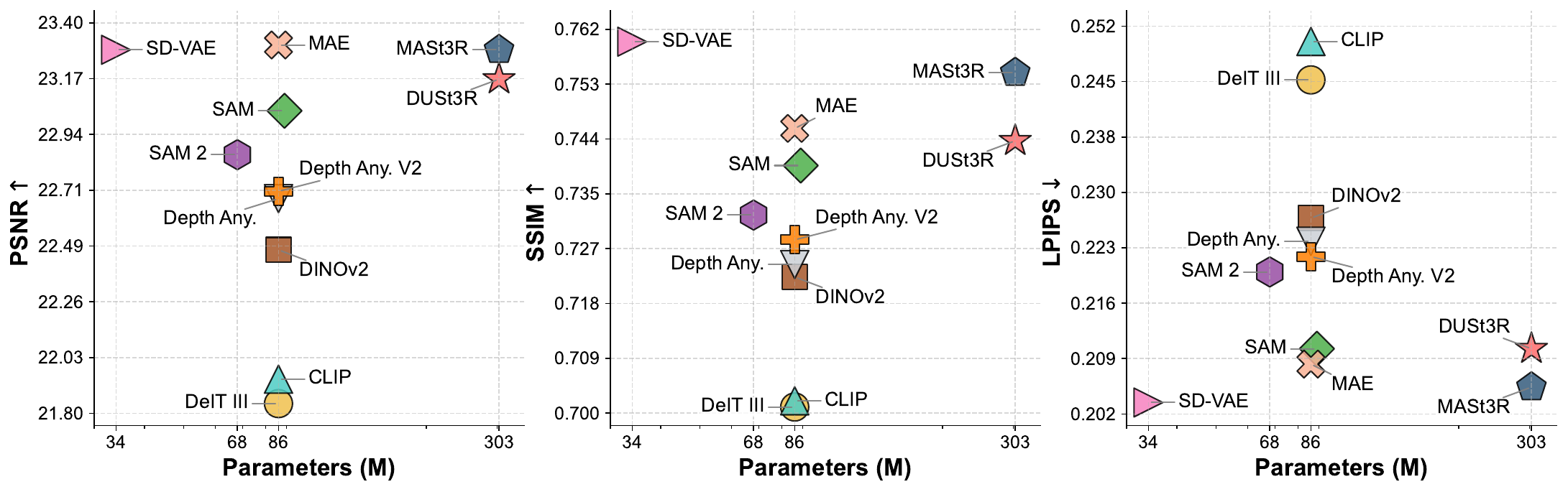}
    \caption{
        \textbf{Performance comparison of various visual foundation models as
            encoders.}
        SD-VAE~\cite{rombach2022high} achieves the best performance with the fewest parameters, demonstrating exceptional effectiveness and efficiency.
        See Appendix~\cref{tab:encoder_performance} for detailed results.
    }
    \label{fig:encoder_performance}
\end{figure*}

\subsection{Ablation Studies}

\noindent\textbf{Visual Encoder.}
We first evaluate various visual foundation models as 3D reconstruction encoders, spanning diverse training objectives, architectures, and scales (see Appendix~\cref{tab:encoder_type}).
As shown in~\cref{fig:encoder_performance}, models trained with pixel-aligned supervision (SD-VAE~\cite{rombach2022high}, MAE~\cite{he2022masked}, MAsT3R~\cite{leroy2025grounding}) systematically outperform those using semantic-aligned supervision (CLIP~\cite{radford2021learning}, DeiT-III~\cite{touvron2022deit}).
SD-VAE~\cite{rombach2022high} achieves optimal performance with superior parameter efficiency while DINOv2~\cite{oquab2023dinov2} substantially underperforms despite its widespread adoption.
This disparity highlights the limitations of semantic-oriented pre-training, which compromises fine-grained spatial information essential for precise geometry and appearance reconstruction.
Consequently, we adopt SD-VAE as our default encoder.

\noindent\textbf{Volumetric Latent Fusion.}
Our latent volume aggregates multi-view correspondences into a unified 3D representation, enabling robust view synthesis across diverse scenes.
As shown in~\cref{tab:plucker_latent}, this component achieves significant gains on RealEstate10K (\gain{+3.99 PSNR, +0.120 SSIM, -0.078 LPIPS}) and demonstrates strong zero-shot generalization: \gain{+4.15 PSNR} on ACID and \gain{+4.98 PSNR} on DTU.
When combined with Plücker coordinates, the latent volume contributes \gain{+3.55 PSNR} on DTU.
These results confirm that volumetric latent fusion effectively learns multi-view correspondences, enhancing reconstruction quality and cross-dataset generalization.

\noindent\textbf{Plücker Coordinates.}
Plücker coordinates provide a compact 6D parameterization of 3D rays, jointly encoding camera origin and viewing direction.
As shown in~\cref{tab:plucker_latent}, incorporating Plücker coordinates yields significant gains on RealEstate10K (\gain{+3.94 PSNR, +0.120 SSIM, -0.077 LPIPS}) and generalizes to unseen datasets: \gain{+3.92 PSNR} on ACID and \gain{+1.41 PSNR} on DTU.
While PSNR improvements become less pronounced when combined with latent volume, which already encodes positional information, Plücker coordinates still enhance perceptual quality on DTU (\gain{+0.025 SSIM, -0.007 LPIPS}).
These results demonstrate that geometric positional encoding complements volumetric correspondences, contributing to reconstruction quality and cross-dataset generalization.

\noindent\textbf{Cost Strategy.}
~\cref{tab:cost_function} evaluates different cost strategies for volumetric latent fusion.
While the difference-based cost computation achieves marginally higher performance on RealEstate10K and ACID, the cost-free approach performs best on the DTU benchmark with \gain{+0.011 SSIM and -0.007 LPIPS} improvements.
These results demonstrate superior generalization across diverse datasets, whereas explicit cost functions exhibit dataset-specific biases.
We adopt the cost-free approach as our default.

\noindent\textbf{Camera-aware Transformer.}
The camera-aware Transformer is essential for ensuring multi-view consistency in novel view synthesis.
As shown in~\cref{tab:ablation}, its removal leads to substantial performance degradation (\gain{-1.07 PSNR, -0.022 SSIM, +0.019 LPIPS}), underscoring its critical role in capturing cross-view relationships.
By incorporating viewpoint-specific information, this module enables the network to maintain geometric coherence across different camera poses, effectively handling occlusions and preserving 3D structural consistency in complex scenes.

\noindent\textbf{Gaussian Decoder.}
The hierarchical Gaussian decoder is essential to the reconstruction pipeline.
As shown in~\cref{tab:ablation}, its inclusion provides a substantial performance improvement (\gain{+2.47 PSNR, +0.072 SSIM, -0.080 LPIPS}), revealing that the progressive upsampling strategy is fundamental to accurate Gaussian parameter prediction.
This hierarchical design decomposes the complex prediction task into progressive refinement stages, facilitating both training stability and multi-scale feature propagation.

\noindent\textbf{Pixel Gradient Loss.}
The pixel gradient loss provides meaningful improvements in reconstruction quality.
As shown in~\cref{tab:ablation}, removing it degrades performance (\gain{-0.12 PSNR, -0.007 SSIM}).
This loss targets high-frequency details and local textures that standard reconstruction losses often overlook, helping preserve fine-grained visual quality.

\noindent\textbf{Convergence Speed.}
Our method demonstrates significant training efficiency.
As shown in~\cref{tab:ablation}, it achieves comparable performance to TranSplat~\cite{zhang2024transplat} in half the training steps (150k vs 300k).
This 2$\times$ speedup stems from the synergy between our latent volume for multi-view fusion, camera-aware Transformer for adaptive correspondence modeling, and hierarchical decoder for progressive refinement.

\begin{table}[!tbp]
    \centering
    \resizebox{\linewidth}{!}{
        \setlength{\tabcolsep}{2pt}
        \begin{tabular}{cc|ccc|ccc|ccc}
            \toprule
            \textbf{Plücker} & \textbf{Latent} & \multicolumn{3}{c|}{\textbf{RealEstate10K}} & \multicolumn{3}{c|}{\textbf{RealEstate10K $\rightarrow$ ACID}} & \multicolumn{3}{c}{\textbf{RealEstate10K $\rightarrow$ DTU}}                                                                                                                                                                   \\
            \textbf{Coord.}  & \textbf{Volume} & \textbf{PSNR$\uparrow$}                     & \textbf{SSIM$\uparrow$}                                        & \textbf{LPIPS$\downarrow$}                                   & \textbf{PSNR$\uparrow$} & \textbf{SSIM$\uparrow$} & \textbf{LPIPS$\downarrow$} & \textbf{PSNR$\uparrow$} & \textbf{SSIM$\uparrow$} & \textbf{LPIPS$\downarrow$} \\
            \midrule
            \xmark           & \xmark          & 22.62                                       & 0.754                                                          & 0.205                                                        & 24.11                   & 0.706                   & 0.242                      & 10.42                   & 0.271                   & 0.604                      \\
            \cmark           & \xmark          & 26.56                                       & 0.874                                                          & 0.128                                                        & 28.03                   & 0.837                   & 0.157                      & 11.83                   & 0.342                   & 0.540                      \\
            \xmark           & \cmark          & 26.61                                       & 0.874                                                          & 0.127                                                        & 28.26                   & \textbf{0.844}          & \textbf{0.150}             & \textbf{15.40}          & 0.562                   & 0.324                      \\
            \rowcolor{heavygray}
            \cmark           & \cmark          & \textbf{26.68}                              & \textbf{0.875}                                                 & \textbf{0.126}                                               & \textbf{28.29}          & \textbf{0.844}          & 0.152                      & 15.38                   & \textbf{0.587}          & \textbf{0.317}             \\
            \bottomrule
        \end{tabular}
    }
    \caption{
        \textbf{Effect of Plücker coordinates and latent volume.}
        Best results in \textbf{bold}; our method in \colorbox{heavygray}{gray}.
    }
    \label{tab:plucker_latent}
\end{table}
\begin{table}[!tbp]
    \centering
    \resizebox{\linewidth}{!}{
        \setlength{\tabcolsep}{2pt}
        \begin{tabular}{c|ccc|ccc|ccc}
            \toprule
            \textbf{Cost}     & \multicolumn{3}{c|}{\textbf{RealEstate10K}} & \multicolumn{3}{c|}{\textbf{RealEstate10K $\to$ ACID}} & \multicolumn{3}{c}{\textbf{RealEstate10K $\to$ DTU}}                                                                                                                                                                   \\
            \textbf{Strategy} & \textbf{PSNR$\uparrow$}                     & \textbf{SSIM$\uparrow$}                                & \textbf{LPIPS$\downarrow$}                           & \textbf{PSNR$\uparrow$} & \textbf{SSIM$\uparrow$} & \textbf{LPIPS$\downarrow$} & \textbf{PSNR$\uparrow$} & \textbf{SSIM$\uparrow$} & \textbf{LPIPS$\downarrow$} \\
            \midrule
            w/o latent volume & 26.56                                       & 0.874                                                  & 0.128                                                & 28.03                   & 0.837                   & 0.157                      & 11.83                   & 0.342                   & 0.540                      \\
            \midrule
            Correlation       & 26.50                                       & 0.872                                                  & 0.129                                                & 28.16                   & 0.842                   & 0.153                      & 15.21                   & 0.567                   & 0.326                      \\
            Difference        & \textbf{26.69}                              & \textbf{0.875}                                         & \textbf{0.126}                                       & \textbf{28.34}          & \textbf{0.844}          & \textbf{0.151}             & 15.36                   & 0.576                   & 0.324                      \\
            \rowcolor{heavygray}
            No cost           & 26.68                                       & \textbf{0.875}                                         & \textbf{0.126}                                       & 28.29                   & \textbf{0.844}          & 0.152                      & \textbf{15.38}          & \textbf{0.587}          & \textbf{0.317}             \\
            \bottomrule
        \end{tabular}
    }
    \caption{
        \textbf{Effect of cost strategies for volumetric latent fusion.}
    }
    \label{tab:cost_function}
\end{table}
\begin{table}[!tbp]
    \centering
    \setlength{\tabcolsep}{2pt}
    \resizebox{\columnwidth}{!}{
        \begin{tabular}{c|c|ccc}
            \toprule
            \textbf{Step}         & \textbf{Design}                      & \textbf{PSNR$\uparrow$}               & \textbf{SSIM$\uparrow$}               & \textbf{LPIPS$\downarrow$}   \\
            \midrule
            \multirow{5}{*}{150k} & \cellcolor{heavygray}{Full}          & \cellcolor{heavygray}{\textbf{26.68}} & \cellcolor{heavygray}{\textbf{0.875}} & \cellcolor{heavygray}{0.126} \\
                                  & w/o camera-aware Transformer         & 25.61                                 & 0.853                                 & 0.145                        \\
                                  & w/o Gaussian decoder                 & 24.21                                 & 0.803                                 & 0.206                        \\
                                  & w/o pixel gradient loss              & 26.56                                 & 0.868                                 & \textbf{0.125}               \\
            \midrule
            \multirow{3}{*}{300k} & MVSplat~\cite{chen2024mvsplat}       & 26.39                                 & 0.869                                 & 0.128                        \\
                                  & eFreeSplat~\cite{min2024epipolar}    & 26.45                                 & 0.865                                 & 0.126                        \\
                                  & TransSplat~\cite{zhang2024transplat} & 26.69                                 & 0.875                                 & 0.125                        \\
            \bottomrule
        \end{tabular}
    }
    \caption{
        \textbf{Ablation studies on RealEstate10K.}
        Our method achieves superior performance with half the training steps.
    }
    \label{tab:ablation}
\end{table}
\section{Conclusion}

We introduced H3R, a hybrid framework that advances generalizable 3D reconstruction by addressing three fundamental challenges: the trade-off in multi-view correspondence modeling, the choice of visual representation and the variability of real-world inputs.
H3R synergistically combines the geometric precision of explicit constraints with the robustness of implicit attention-based aggregation.
By integrating a cost-free latent volume with a camera-aware Transformer, our model achieves both geometric precision and robust correspondence, converging $2\times$ faster than prior art.
Furthermore, we identified a critical mismatch between feature representation and reconstruction fidelity.
Our systematic analysis reveals that spatial-aligned foundation models (\textit{e.g.}, SD-VAE) significantly outperform widely adopted semantic-aligned models like DINOv2.
This provides a crucial insight for future work: prioritizing spatial fidelity over semantic abstraction is key to achieving high-fidelity 3D reconstruction.
Experimental results confirm our method's efficacy, demonstrating consistent improvements across multiple benchmarks and robust cross-dataset generalization while supporting variable-number or high-resolution input views.
H3R not only delivers superior performance but also provides foundational principles for designing more efficient and robust reconstruction systems.

\clearpage
\section{Acknowledgments}
This work is partially supported by the National Science and Technology Major Project (Grant No.
2022ZD0117802), the Fundamental Research Funds for the Central Universities (Grant No. 226-2025-00055), and the Earth System Big Data Platform of the School of Earth Sciences, Zhejiang University.
This research is also supported by the Agency for Science, Technology and Research (A*STAR) under its MTC Programmatic Funds (Grant No.
M23L7b0021).

{
    \small
    \bibliographystyle{ieeenat_fullname}
    \bibliography{main}

\begin{thebibliography}{89}
\providecommand{\natexlab}[1]{#1}
\providecommand{\url}[1]{\texttt{#1}}
\expandafter\ifx\csname urlstyle\endcsname\relax
  \providecommand{\doi}[1]{doi: #1}\else
  \providecommand{\doi}{doi: \begingroup \urlstyle{rm}\Url}\fi

\bibitem[Cao et~al.(2024)Cao, Ren, and Fu]{cao2024mvsformer++}
Chenjie Cao, Xinlin Ren, and Yanwei Fu.
\newblock Mvsformer++: Revealing the devil in transformer's details for multi-view stereo.
\newblock In \emph{ICLR}, 2024.

\bibitem[Caron et~al.(2021)Caron, Touvron, Misra, J{\'e}gou, Mairal, Bojanowski, and Joulin]{caron2021emerging}
Mathilde Caron, Hugo Touvron, Ishan Misra, Herv{\'e} J{\'e}gou, Julien Mairal, Piotr Bojanowski, and Armand Joulin.
\newblock Emerging properties in self-supervised vision transformers.
\newblock In \emph{ICCV}, 2021.

\bibitem[Chan et~al.(2023)Chan, Nagano, Chan, Bergman, Park, Levy, Aittala, De~Mello, Karras, and Wetzstein]{chan2023generative}
Eric~R Chan, Koki Nagano, Matthew~A Chan, Alexander~W Bergman, Jeong~Joon Park, Axel Levy, Miika Aittala, Shalini De~Mello, Tero Karras, and Gordon Wetzstein.
\newblock Generative novel view synthesis with 3d-aware diffusion models.
\newblock In \emph{ICCV}, 2023.

\bibitem[Charatan et~al.(2024)Charatan, Li, Tagliasacchi, and Sitzmann]{charatan2024pixelsplat}
David Charatan, Sizhe~Lester Li, Andrea Tagliasacchi, and Vincent Sitzmann.
\newblock pixelsplat: 3d gaussian splats from image pairs for scalable generalizable 3d reconstruction.
\newblock In \emph{CVPR}, 2024.

\bibitem[Chen et~al.(2021)Chen, Xu, Zhao, Zhang, Xiang, Yu, and Su]{chen2021mvsnerf}
Anpei Chen, Zexiang Xu, Fuqiang Zhao, Xiaoshuai Zhang, Fanbo Xiang, Jingyi Yu, and Hao Su.
\newblock Mvsnerf: Fast generalizable radiance field reconstruction from multi-view stereo.
\newblock In \emph{ICCV}, 2021.

\bibitem[Chen et~al.(2024{\natexlab{a}})Chen, Xu, Esposito, Tang, and Geiger]{chen2025lara}
Anpei Chen, Haofei Xu, Stefano Esposito, Siyu Tang, and Andreas Geiger.
\newblock Lara: Efficient large-baseline radiance fields.
\newblock In \emph{ECCV}, 2024{\natexlab{a}}.

\bibitem[Chen et~al.(2024{\natexlab{b}})Chen, Xu, Zheng, Zhuang, Pollefeys, Geiger, Cham, and Cai]{chen2024mvsplat}
Yuedong Chen, Haofei Xu, Chuanxia Zheng, Bohan Zhuang, Marc Pollefeys, Andreas Geiger, Tat-Jen Cham, and Jianfei Cai.
\newblock Mvsplat: Efficient 3d gaussian splatting from sparse multi-view images.
\newblock In \emph{ECCV}, 2024{\natexlab{b}}.

\bibitem[Chen et~al.(2024{\natexlab{c}})Chen, Zheng, Xu, Zhuang, Vedaldi, Cham, and Cai]{chen2024mvsplat360}
Yuedong Chen, Chuanxia Zheng, Haofei Xu, Bohan Zhuang, Andrea Vedaldi, Tat-Jen Cham, and Jianfei Cai.
\newblock Mvsplat360: Feed-forward 360 scene synthesis from sparse views.
\newblock In \emph{NeurIPS}, 2024{\natexlab{c}}.

\bibitem[Collins(1996)]{collins1996space}
Robert~T Collins.
\newblock A space-sweep approach to true multi-image matching.
\newblock In \emph{CVPR}, 1996.

\bibitem[Cong et~al.(2023)Cong, Liang, Wang, Fan, Chen, Varma, Wang, and Wang]{cong2023enhancing}
Wenyan Cong, Hanxue Liang, Peihao Wang, Zhiwen Fan, Tianlong Chen, Mukund Varma, Yi Wang, and Zhangyang Wang.
\newblock Enhancing nerf akin to enhancing llms: Generalizable nerf transformer with mixture-of-view-experts.
\newblock In \emph{ICCV}, 2023.

\bibitem[Deitke et~al.(2023{\natexlab{a}})Deitke, Liu, Wallingford, Ngo, Michel, Kusupati, Fan, Laforte, Voleti, Gadre, et~al.]{deitke2023objaversexl}
Matt Deitke, Ruoshi Liu, Matthew Wallingford, Huong Ngo, Oscar Michel, Aditya Kusupati, Alan Fan, Christian Laforte, Vikram Voleti, Samir~Yitzhak Gadre, et~al.
\newblock Objaverse-xl: A universe of 10m+ 3d objects.
\newblock In \emph{NeurIPS}, 2023{\natexlab{a}}.

\bibitem[Deitke et~al.(2023{\natexlab{b}})Deitke, Schwenk, Salvador, Weihs, Michel, VanderBilt, Schmidt, Ehsani, Kembhavi, and Farhadi]{deitke2023objaverse}
Matt Deitke, Dustin Schwenk, Jordi Salvador, Luca Weihs, Oscar Michel, Eli VanderBilt, Ludwig Schmidt, Kiana Ehsani, Aniruddha Kembhavi, and Ali Farhadi.
\newblock Objaverse: A universe of annotated 3d objects.
\newblock In \emph{CVPR}, 2023{\natexlab{b}}.

\bibitem[Ding et~al.(2022)Ding, Yuan, Zhu, Zhang, Liu, Wang, and Liu]{ding2022transmvsnet}
Yikang Ding, Wentao Yuan, Qingtian Zhu, Haotian Zhang, Xiangyue Liu, Yuanjiang Wang, and Xiao Liu.
\newblock Transmvsnet: Global context-aware multi-view stereo network with transformers.
\newblock In \emph{CVPR}, 2022.

\bibitem[Du et~al.(2023)Du, Smith, Tewari, and Sitzmann]{du2023learning}
Yilun Du, Cameron Smith, Ayush Tewari, and Vincent Sitzmann.
\newblock Learning to render novel views from wide-baseline stereo pairs.
\newblock In \emph{CVPR}, 2023.

\bibitem[Fei et~al.(2024)Fei, Zheng, Duan, Zhan, Tomizuka, Keutzer, and Lu]{fei2024pixelgaussian}
Xin Fei, Wenzhao Zheng, Yueqi Duan, Wei Zhan, Masayoshi Tomizuka, Kurt Keutzer, and Jiwen Lu.
\newblock Pixelgaussian: Generalizable 3d gaussian reconstruction from arbitrary views.
\newblock \emph{arXiv:2410.18979}, 2024.

\bibitem[Gu et~al.(2020)Gu, Fan, Zhu, Dai, Tan, and Tan]{gu2020casmvsnet}
Xiaodong Gu, Zhiwen Fan, Siyu Zhu, Zuozhuo Dai, Feitong Tan, and Ping Tan.
\newblock Cascade cost volume for high-resolution multi-view stereo and stereo matching.
\newblock In \emph{CVPR}, 2020.

\bibitem[He et~al.(2016)He, Zhang, Ren, and Sun]{he2016deep}
Kaiming He, Xiangyu Zhang, Shaoqing Ren, and Jian Sun.
\newblock Deep residual learning for image recognition.
\newblock In \emph{CVPR}, 2016.

\bibitem[He et~al.(2022)He, Chen, Xie, Li, Doll{\'a}r, and Girshick]{he2022masked}
Kaiming He, Xinlei Chen, Saining Xie, Yanghao Li, Piotr Doll{\'a}r, and Ross Girshick.
\newblock Masked autoencoders are scalable vision learners.
\newblock In \emph{CVPR}, 2022.

\bibitem[Henry et~al.(2020)Henry, Dachapally, Pawar, and Chen]{henry2020query}
Alex Henry, Prudhvi~Raj Dachapally, Shubham Pawar, and Yuxuan Chen.
\newblock Query-key normalization for transformers.
\newblock In \emph{EMNLP}, 2020.

\bibitem[Hoffmann et~al.(2022)Hoffmann, Borgeaud, Mensch, Buchatskaya, Cai, Rutherford, de~Las~Casas, Hendricks, Welbl, Clark, et~al.]{hoffmann2022training}
Jordan Hoffmann, Sebastian Borgeaud, Arthur Mensch, Elena Buchatskaya, Trevor Cai, Eliza Rutherford, Diego de Las~Casas, Lisa~Anne Hendricks, Johannes Welbl, Aidan Clark, et~al.
\newblock Training compute-optimal large language models.
\newblock In \emph{NeurIPS}, 2022.

\bibitem[Hong et~al.(2023)Hong, Zhang, Gu, Bi, Zhou, Liu, Liu, Sunkavalli, Bui, and Tan]{hong2023lrm}
Yicong Hong, Kai Zhang, Jiuxiang Gu, Sai Bi, Yang Zhou, Difan Liu, Feng Liu, Kalyan Sunkavalli, Trung Bui, and Hao Tan.
\newblock Lrm: Large reconstruction model for single image to 3d.
\newblock In \emph{ICLR}, 2023.

\bibitem[Im et~al.(2019)Im, Jeon, Lin, and Kweon]{im2019dpsnet}
Sunghoon Im, Hae-Gon Jeon, Stephen Lin, and In~So Kweon.
\newblock Dpsnet: End-to-end deep plane sweep stereo.
\newblock In \emph{ICLR}, 2019.

\bibitem[Jena et~al.(2025)Jena, Vutukur, and Boukhayma]{jena2025sparsplat}
Shubhendu Jena, Shishir~Reddy Vutukur, and Adnane Boukhayma.
\newblock Sparsplat: Fast multi-view reconstruction with generalizable 2d gaussian splatting.
\newblock In \emph{CVPR Workshops}, 2025.

\bibitem[Jensen et~al.(2014)Jensen, Dahl, Vogiatzis, Tola, and Aan{\ae}s]{jensen2014large}
Rasmus Jensen, Anders Dahl, George Vogiatzis, Engin Tola, and Henrik Aan{\ae}s.
\newblock Large scale multi-view stereopsis evaluation.
\newblock In \emph{CVPR}, 2014.

\bibitem[Jiang et~al.(2023)Jiang, Jiang, Zhao, and Huang]{jiang2023leap}
Hanwen Jiang, Zhenyu Jiang, Yue Zhao, and Qixing Huang.
\newblock Leap: Liberate sparse-view 3d modeling from camera poses.
\newblock In \emph{ICLR}, 2023.

\bibitem[Jiang et~al.(2025)Jiang, Zhao, Rahmani, Soh, Liu, and Zhao]{jianggaussianblock}
Shuyi Jiang, Qihao Zhao, Hossein Rahmani, De~Wen Soh, Jun Liu, and Na Zhao.
\newblock Gaussianblock: Building part-aware compositional and editable 3d scene by primitives and gaussians.
\newblock In \emph{ICLR}, 2025.

\bibitem[Jin et~al.(2024)Jin, Jiang, Tan, Zhang, Bi, Zhang, Luan, Snavely, and Xu]{jin2024lvsm}
Haian Jin, Hanwen Jiang, Hao Tan, Kai Zhang, Sai Bi, Tianyuan Zhang, Fujun Luan, Noah Snavely, and Zexiang Xu.
\newblock Lvsm: A large view synthesis model with minimal 3d inductive bias.
\newblock In \emph{ICLR}, 2024.

\bibitem[Johari et~al.(2022)Johari, Lepoittevin, and Fleuret]{johari2022geonerf}
Mohammad~Mahdi Johari, Yann Lepoittevin, and Fran{\c{c}}ois Fleuret.
\newblock Geonerf: Generalizing nerf with geometry priors.
\newblock In \emph{CVPR}, 2022.

\bibitem[Kaplan et~al.(2020)Kaplan, McCandlish, Henighan, Brown, Chess, Child, Gray, Radford, Wu, and Amodei]{kaplan2020scaling}
Jared Kaplan, Sam McCandlish, Tom Henighan, Tom~B Brown, Benjamin Chess, Rewon Child, Scott Gray, Alec Radford, Jeffrey Wu, and Dario Amodei.
\newblock Scaling laws for neural language models.
\newblock \emph{arXiv:2001.08361}, 2020.

\bibitem[Kerbl et~al.(2023)Kerbl, Kopanas, Leimk{\"u}hler, and Drettakis]{kerbl20233d}
Bernhard Kerbl, Georgios Kopanas, Thomas Leimk{\"u}hler, and George Drettakis.
\newblock 3d gaussian splatting for real-time radiance field rendering.
\newblock \emph{ACM TOG}, 2023.

\bibitem[Kirillov et~al.(2023)Kirillov, Mintun, Ravi, Mao, Rolland, Gustafson, Xiao, Whitehead, Berg, Lo, et~al.]{kirillov2023segment}
Alexander Kirillov, Eric Mintun, Nikhila Ravi, Hanzi Mao, Chlo{\'{e}} Rolland, Laura Gustafson, Tete Xiao, Spencer Whitehead, Alexander~C. Berg, Wan{-}Yen Lo, et~al.
\newblock Segment anything.
\newblock In \emph{ICCV}, 2023.

\bibitem[Kulhanek et~al.(2024)Kulhanek, Peng, Kukelova, Pollefeys, and Sattler]{kulhanek2024wildgaussians}
Jonas Kulhanek, Songyou Peng, Zuzana Kukelova, Marc Pollefeys, and Torsten Sattler.
\newblock Wildgaussians: 3d gaussian splatting in the wild.
\newblock In \emph{NeurIPS}, 2024.

\bibitem[Leroy et~al.(2025)Leroy, Cabon, and Revaud]{leroy2025grounding}
Vincent Leroy, Yohann Cabon, and J{\'e}r{\^o}me Revaud.
\newblock Grounding image matching in 3d with mast3r.
\newblock In \emph{ECCV}, 2025.

\bibitem[Li et~al.(2024)Li, Zhang, Dai, Liu, Cheng, Li, Wang, and Han]{li2024gp}
Hao Li, Dingwen Zhang, Yalun Dai, Nian Liu, Lechao Cheng, Jingfeng Li, Jingdong Wang, and Junwei Han.
\newblock Gp-nerf: Generalized perception nerf for context-aware 3d scene understanding.
\newblock In \emph{CVPR}, 2024.

\bibitem[Liu et~al.(2021)Liu, Tucker, Jampani, Makadia, Snavely, and Kanazawa]{liu2021infinite}
Andrew Liu, Richard Tucker, Varun Jampani, Ameesh Makadia, Noah Snavely, and Angjoo Kanazawa.
\newblock Infinite nature: Perpetual view generation of natural scenes from a single image.
\newblock In \emph{ICCV}, 2021.

\bibitem[Liu et~al.(2023{\natexlab{a}})Liu, Zhang, Zheng, and Duan]{liu2023semantic}
Fangfu Liu, Chubin Zhang, Yu Zheng, and Yueqi Duan.
\newblock Semantic ray: Learning a generalizable semantic field with cross-reprojection attention.
\newblock In \emph{CVPR}, 2023{\natexlab{a}}.

\bibitem[Liu et~al.(2023{\natexlab{b}})Liu, Wu, Van~Hoorick, Tokmakov, Zakharov, and Vondrick]{liu2023zero}
Ruoshi Liu, Rundi Wu, Basile Van~Hoorick, Pavel Tokmakov, Sergey Zakharov, and Carl Vondrick.
\newblock Zero-1-to-3: Zero-shot one image to 3d object.
\newblock In \emph{ICCV}, 2023{\natexlab{b}}.

\bibitem[Liu et~al.(2023{\natexlab{c}})Liu, Ye, Zhao, Pan, Shi, and Cao]{liu2023epipolar}
Tianqi Liu, Xinyi Ye, Weiyue Zhao, Zhiyu Pan, Min Shi, and Zhiguo Cao.
\newblock When epipolar constraint meets non-local operators in multi-view stereo.
\newblock In \emph{ICCV}, 2023{\natexlab{c}}.

\bibitem[Liu et~al.(2024)Liu, Wang, Hu, Shen, Ye, Zang, Cao, Li, and Liu]{liu2024mvsgaussian}
Tianqi Liu, Guangcong Wang, Shoukang Hu, Liao Shen, Xinyi Ye, Yuhang Zang, Zhiguo Cao, Wei Li, and Ziwei Liu.
\newblock Mvsgaussian: Fast generalizable gaussian splatting reconstruction from multi-view stereo.
\newblock In \emph{ECCV}, 2024.

\bibitem[Liu et~al.(2022)Liu, Peng, Liu, Wang, Wang, Theobalt, Zhou, and Wang]{liu2022neural}
Yuan Liu, Sida Peng, Lingjie Liu, Qianqian Wang, Peng Wang, Christian Theobalt, Xiaowei Zhou, and Wenping Wang.
\newblock Neural rays for occlusion-aware image-based rendering.
\newblock In \emph{CVPR}, 2022.

\bibitem[Mildenhall et~al.(2019)Mildenhall, Srinivasan, Ortiz-Cayon, Kalantari, Ramamoorthi, Ng, and Kar]{mildenhall2019local}
Ben Mildenhall, Pratul~P Srinivasan, Rodrigo Ortiz-Cayon, Nima~Khademi Kalantari, Ravi Ramamoorthi, Ren Ng, and Abhishek Kar.
\newblock Local light field fusion: Practical view synthesis with prescriptive sampling guidelines.
\newblock \emph{ACM TOG}, 2019.

\bibitem[Min et~al.(2024)Min, Luo, Sun, and Yang]{min2024epipolar}
Zhiyuan Min, Yawei Luo, Jianwen Sun, and Yi Yang.
\newblock Epipolar-free 3d gaussian splatting for generalizable novel view synthesis.
\newblock In \emph{NeurIPS}, 2024.

\bibitem[Oquab et~al.(2023)Oquab, Darcet, Moutakanni, Vo, Szafraniec, Khalidov, Fernandez, Haziza, Massa, El{-}Nouby, et~al.]{oquab2023dinov2}
Maxime Oquab, Timoth{\'{e}}e Darcet, Th{\'{e}}o Moutakanni, Huy~V. Vo, Marc Szafraniec, Vasil Khalidov, Pierre Fernandez, Daniel Haziza, Francisco Massa, Alaaeldin El{-}Nouby, et~al.
\newblock Dinov2: Learning robust visual features without supervision.
\newblock \emph{TMLR}, 2023.

\bibitem[Peng et~al.(2022)Peng, Wang, Wang, Lai, and Wang]{peng2022unimvsnet}
Rui Peng, Rongjie Wang, Zhenyu Wang, Yawen Lai, and Ronggang Wang.
\newblock Rethinking depth estimation for multi-view stereo: A unified representation.
\newblock In \emph{CVPR}, 2022.

\bibitem[Qiu et~al.(2024)Qiu, Chen, Gu, Zuo, Xu, Wu, Yuan, Dong, Bo, and Han]{qiu2024richdreamer}
Lingteng Qiu, Guanying Chen, Xiaodong Gu, Qi Zuo, Mutian Xu, Yushuang Wu, Weihao Yuan, Zilong Dong, Liefeng Bo, and Xiaoguang Han.
\newblock Richdreamer: A generalizable normal-depth diffusion model for detail richness in text-to-3d.
\newblock In \emph{CVPR}, 2024.

\bibitem[Radford et~al.(2021)Radford, Kim, Hallacy, Ramesh, Goh, Agarwal, Sastry, Askell, Mishkin, Clark, et~al.]{radford2021learning}
Alec Radford, Jong~Wook Kim, Chris Hallacy, Aditya Ramesh, Gabriel Goh, Sandhini Agarwal, Girish Sastry, Amanda Askell, Pamela Mishkin, Jack Clark, et~al.
\newblock Learning transferable visual models from natural language supervision.
\newblock In \emph{ICML}, 2021.

\bibitem[Ravi et~al.(2024)Ravi, Gabeur, Hu, Hu, Ryali, Ma, Khedr, R{\"{a}}dle, Rolland, Gustafson, et~al.]{ravi2024sam}
Nikhila Ravi, Valentin Gabeur, Yuan{-}Ting Hu, Ronghang Hu, Chaitanya Ryali, Tengyu Ma, Haitham Khedr, Roman R{\"{a}}dle, Chlo{\'{e}} Rolland, Laura Gustafson, et~al.
\newblock Sam 2: Segment anything in images and videos.
\newblock In \emph{ICLR}, 2024.

\bibitem[Rombach et~al.(2022)Rombach, Blattmann, Lorenz, Esser, and Ommer]{rombach2022high}
Robin Rombach, Andreas Blattmann, Dominik Lorenz, Patrick Esser, and Bj{\"o}rn Ommer.
\newblock High-resolution image synthesis with latent diffusion models.
\newblock In \emph{CVPR}, 2022.

\bibitem[Ryali et~al.(2023)Ryali, Hu, Bolya, Wei, Fan, Huang, Aggarwal, Chowdhury, Poursaeed, Hoffman, Malik, Li, and Feichtenhofer]{ryali2023hiera}
Chaitanya Ryali, Yuan{-}Ting Hu, Daniel Bolya, Chen Wei, Haoqi Fan, Po{-}Yao Huang, Vaibhav Aggarwal, Arkabandhu Chowdhury, Omid Poursaeed, Judy Hoffman, Jitendra Malik, Yanghao Li, and Christoph Feichtenhofer.
\newblock Hiera: A hierarchical vision transformer without the bells-and-whistles.
\newblock In \emph{ICML}, 2023.

\bibitem[Sajjadi et~al.(2022)Sajjadi, Meyer, Pot, Bergmann, Greff, Radwan, Vora, Lucic, Duckworth, Dosovitskiy, et~al.]{sajjadi2022scene}
Mehdi S.~M. Sajjadi, Henning Meyer, Etienne Pot, Urs Bergmann, Klaus Greff, Noha Radwan, Suhani Vora, Mario Lucic, Daniel Duckworth, Alexey Dosovitskiy, et~al.
\newblock Scene representation transformer: Geometry-free novel view synthesis through set-latent scene representations.
\newblock In \emph{CVPR}, 2022.

\bibitem[Sch\"{o}nberger and Frahm(2016)]{schoenberger2016sfm}
Johannes~Lutz Sch\"{o}nberger and Jan-Michael Frahm.
\newblock Structure-from-motion revisited.
\newblock In \emph{CVPR}, 2016.

\bibitem[Shazeer(2020)]{shazeer2020glu}
Noam Shazeer.
\newblock Glu variants improve transformer.
\newblock \emph{arXiv:2002.05202}, 2020.

\bibitem[Shen et~al.(2025)Shen, Wu, Yi, Zhou, Zhang, Yan, and Wang]{shen2025gamba}
Qiuhong Shen, Zike Wu, Xuanyu Yi, Pan Zhou, Hanwang Zhang, Shuicheng Yan, and Xinchao Wang.
\newblock Gamba: Marry gaussian splatting with mamba for single-view 3d reconstruction.
\newblock \emph{IEEE TPAMI}, 2025.

\bibitem[Shi et~al.(2023)Shi, Wang, Ye, Long, Li, and Yang]{shi2023mvdream}
Yichun Shi, Peng Wang, Jianglong Ye, Mai Long, Kejie Li, and Xiao Yang.
\newblock Mvdream: Multi-view diffusion for 3d generation.
\newblock In \emph{ICLR}, 2023.

\bibitem[society~of London(1864)]{royal1864philosophical}
Royal society~of London.
\newblock \emph{Philosophical Transactions of the Royal Society of London}.
\newblock 1864.

\bibitem[Suhail et~al.(2022)Suhail, Esteves, Sigal, and Makadia]{suhail2022generalizable}
Mohammed Suhail, Carlos Esteves, Leonid Sigal, and Ameesh Makadia.
\newblock Generalizable patch-based neural rendering.
\newblock In \emph{ECCV}, 2022.

\bibitem[Szymanowicz et~al.(2024)Szymanowicz, Rupprecht, and Vedaldi]{szymanowicz2024splatter}
Stanislaw Szymanowicz, Chrisitian Rupprecht, and Andrea Vedaldi.
\newblock Splatter image: Ultra-fast single-view 3d reconstruction.
\newblock In \emph{CVPR}, 2024.

\bibitem[Tang et~al.(2024{\natexlab{a}})Tang, Chen, Chen, Wang, Zeng, and Liu]{tang2025lgm}
Jiaxiang Tang, Zhaoxi Chen, Xiaokang Chen, Tengfei Wang, Gang Zeng, and Ziwei Liu.
\newblock Lgm: Large multi-view gaussian model for high-resolution 3d content creation.
\newblock In \emph{ECCV}, 2024{\natexlab{a}}.

\bibitem[Tang et~al.(2024{\natexlab{b}})Tang, Ye, Ye, Lin, Zhou, Chen, and Ouyang]{tang2024hisplat}
Shengji Tang, Weicai Ye, Peng Ye, Weihao Lin, Yang Zhou, Tao Chen, and Wanli Ouyang.
\newblock Hisplat: Hierarchical 3d gaussian splatting for generalizable sparse-view reconstruction.
\newblock \emph{ICLR}, 2024{\natexlab{b}}.

\bibitem[Touvron et~al.(2022)Touvron, Cord, and J{\'e}gou]{touvron2022deit}
Hugo Touvron, Matthieu Cord, and Herv{\'e} J{\'e}gou.
\newblock Deit iii: Revenge of the vit.
\newblock In \emph{ECCV}, 2022.

\bibitem[Touvron et~al.(2023)Touvron, Lavril, Izacard, Martinet, Lachaux, Lacroix, Rozi{\`{e}}re, Goyal, Hambro, Azhar, et~al.]{touvron2023llama}
Hugo Touvron, Thibaut Lavril, Gautier Izacard, Xavier Martinet, Marie{-}Anne Lachaux, Timoth{\'{e}}e Lacroix, Baptiste Rozi{\`{e}}re, Naman Goyal, Eric Hambro, Faisal Azhar, et~al.
\newblock Llama: Open and efficient foundation language models.
\newblock \emph{arXiv:2302.13971}, 2023.

\bibitem[Trevithick and Yang(2021)]{trevithick2021grf}
Alex Trevithick and Bo Yang.
\newblock Grf: Learning a general radiance field for 3d representation and rendering.
\newblock In \emph{ICCV}, 2021.

\bibitem[Wang and Zhao(2024)]{wang2024gs2}
Chengshun Wang and Na Zhao.
\newblock Gs2-gnesf: Geometry-semantics synergy for generalizable neural semantic fields.
\newblock In \emph{ACM MM}, 2024.

\bibitem[Wang et~al.(2021{\natexlab{a}})Wang, Liu, Liu, Theobalt, Komura, and Wang]{wang2021neus}
Peng Wang, Lingjie Liu, Yuan Liu, Christian Theobalt, Taku Komura, and Wenping Wang.
\newblock Neus: Learning neural implicit surfaces by volume rendering for multi-view reconstruction.
\newblock In \emph{NeurIPS}, 2021{\natexlab{a}}.

\bibitem[Wang et~al.(2021{\natexlab{b}})Wang, Wang, Genova, Srinivasan, Zhou, Barron, Martin-Brualla, Snavely, and Funkhouser]{wang2021ibrnet}
Qianqian Wang, Zhicheng Wang, Kyle Genova, Pratul~P Srinivasan, Howard Zhou, Jonathan~T Barron, Ricardo Martin-Brualla, Noah Snavely, and Thomas Funkhouser.
\newblock Ibrnet: Learning multi-view image-based rendering.
\newblock In \emph{CVPR}, 2021{\natexlab{b}}.

\bibitem[Wang et~al.(2024{\natexlab{a}})Wang, Leroy, Cabon, Chidlovskii, and Revaud]{wang2024dust3r}
Shuzhe Wang, Vincent Leroy, Yohann Cabon, Boris Chidlovskii, and Jerome Revaud.
\newblock Dust3r: Geometric 3d vision made easy.
\newblock In \emph{CVPR}, 2024{\natexlab{a}}.

\bibitem[Wang et~al.(2021{\natexlab{c}})Wang, Xie, Dong, and Shan]{wang2021real}
Xintao Wang, Liangbin Xie, Chao Dong, and Ying Shan.
\newblock Real-esrgan: Training real-world blind super-resolution with pure synthetic data.
\newblock In \emph{ICCV}, 2021{\natexlab{c}}.

\bibitem[Wang et~al.(2024{\natexlab{b}})Wang, Huang, Chen, and Lee]{wang2024freesplat}
Yunsong Wang, Tianxin Huang, Hanlin Chen, and Gim~Hee Lee.
\newblock Freesplat: Generalizable 3d gaussian splatting towards free view synthesis of indoor scenes.
\newblock \emph{NeurIPS}, 2024{\natexlab{b}}.

\bibitem[Wang et~al.(2024{\natexlab{c}})Wang, Yi, Wu, Zhao, Chen, and Zhang]{wang2024view}
Yuxuan Wang, Xuanyu Yi, Zike Wu, Na Zhao, Long Chen, and Hanwang Zhang.
\newblock View-consistent 3d editing with gaussian splatting.
\newblock In \emph{ECCV}, 2024{\natexlab{c}}.

\bibitem[Wang et~al.(2004)Wang, Bovik, Sheikh, and Simoncelli]{wang2004image}
Zhou Wang, Alan~C Bovik, Hamid~R Sheikh, and Eero~P Simoncelli.
\newblock Image quality assessment: from error visibility to structural similarity.
\newblock \emph{IEEE TIP}, 2004.

\bibitem[Wewer et~al.(2024)Wewer, Raj, Ilg, Schiele, and Lenssen]{wewer2025latentsplat}
Christopher Wewer, Kevin Raj, Eddy Ilg, Bernt Schiele, and Jan~Eric Lenssen.
\newblock latentsplat: Autoencoding variational gaussians for fast generalizable 3d reconstruction.
\newblock In \emph{ECCV}, 2024.

\bibitem[Xie et~al.(2024)Xie, Bi, Shu, Zhang, Xu, Zhou, Pirk, Kaufman, Sun, and Tan]{xie2024lrm}
Desai Xie, Sai Bi, Zhixin Shu, Kai Zhang, Zexiang Xu, Yi Zhou, S{\"o}ren Pirk, Arie Kaufman, Xin Sun, and Hao Tan.
\newblock Lrm-zero: Training large reconstruction models with synthesized data.
\newblock \emph{NeurIPS}, 2024.

\bibitem[Xiong et~al.(2023)Xiong, Peng, Zhang, Feng, Jiao, Gao, and Wang]{xiong2023cl}
Kaiqiang Xiong, Rui Peng, Zhe Zhang, Tianxing Feng, Jianbo Jiao, Feng Gao, and Ronggang Wang.
\newblock Cl-mvsnet: Unsupervised multi-view stereo with dual-level contrastive learning.
\newblock In \emph{ICCV}, 2023.

\bibitem[Xu et~al.(2023{\natexlab{a}})Xu, Zhang, Cai, Rezatofighi, Yu, Tao, and Geiger]{xu2023unifying}
Haofei Xu, Jing Zhang, Jianfei Cai, Hamid Rezatofighi, Fisher Yu, Dacheng Tao, and Andreas Geiger.
\newblock Unifying flow, stereo and depth estimation.
\newblock \emph{ICCV Workshops}, 2023{\natexlab{a}}.

\bibitem[Xu et~al.(2024{\natexlab{a}})Xu, Chen, Chen, Sakaridis, Zhang, Pollefeys, Geiger, and Yu]{xu2024murf}
Haofei Xu, Anpei Chen, Yuedong Chen, Christos Sakaridis, Yulun Zhang, Marc Pollefeys, Andreas Geiger, and Fisher Yu.
\newblock Murf: Multi-baseline radiance fields.
\newblock In \emph{CVPR}, 2024{\natexlab{a}}.

\bibitem[Xu et~al.(2024{\natexlab{b}})Xu, Peng, Wang, Blum, Barath, Geiger, and Pollefeys]{xu2024depthsplat}
Haofei Xu, Songyou Peng, Fangjinhua Wang, Hermann Blum, Daniel Barath, Andreas Geiger, and Marc Pollefeys.
\newblock Depthsplat: Connecting gaussian splatting and depth.
\newblock \emph{CVPR}, 2024{\natexlab{b}}.

\bibitem[Xu et~al.(2023{\natexlab{b}})Xu, Tan, Luan, Bi, Wang, Li, Shi, Sunkavalli, Wetzstein, Xu, and Zhang]{xu2023dmv3d}
Yinghao Xu, Hao Tan, Fujun Luan, Sai Bi, Peng Wang, Jiahao Li, Zifan Shi, Kalyan Sunkavalli, Gordon Wetzstein, Zexiang Xu, and Kai Zhang.
\newblock Dmv3d: Denoising multi-view diffusion using 3d large reconstruction model.
\newblock In \emph{ICLR}, 2023{\natexlab{b}}.

\bibitem[Xu et~al.(2024{\natexlab{c}})Xu, Shi, Yifan, Chen, Yang, Peng, Shen, and Wetzstein]{xu2024grm}
Yinghao Xu, Zifan Shi, Wang Yifan, Hansheng Chen, Ceyuan Yang, Sida Peng, Yujun Shen, and Gordon Wetzstein.
\newblock Grm: Large gaussian reconstruction model for efficient 3d reconstruction and generation.
\newblock In \emph{ECCV}, 2024{\natexlab{c}}.

\bibitem[Yang et~al.(2024{\natexlab{a}})Yang, Kang, Huang, Xu, Feng, and Zhao]{yang2024depth}
Lihe Yang, Bingyi Kang, Zilong Huang, Xiaogang Xu, Jiashi Feng, and Hengshuang Zhao.
\newblock Depth anything: Unleashing the power of large-scale unlabeled data.
\newblock In \emph{CVPR}, 2024{\natexlab{a}}.

\bibitem[Yang et~al.(2024{\natexlab{b}})Yang, Kang, Huang, Zhao, Xu, Feng, and Zhao]{yang2024depthv2}
Lihe Yang, Bingyi Kang, Zilong Huang, Zhen Zhao, Xiaogang Xu, Jiashi Feng, and Hengshuang Zhao.
\newblock Depth anything v2.
\newblock \emph{NeurIPS}, 2024{\natexlab{b}}.

\bibitem[Yao et~al.(2018)Yao, Luo, Li, Fang, and Quan]{yao2018mvsnet}
Yao Yao, Zixin Luo, Shiwei Li, Tian Fang, and Long Quan.
\newblock Mvsnet: Depth inference for unstructured multi-view stereo.
\newblock In \emph{ECCV}, 2018.

\bibitem[Yu et~al.(2021)Yu, Ye, Tancik, and Kanazawa]{yu2021pixelnerf}
Alex Yu, Vickie Ye, Matthew Tancik, and Angjoo Kanazawa.
\newblock pixelnerf: Neural radiance fields from one or few images.
\newblock In \emph{CVPR}, 2021.

\bibitem[Zhang et~al.(2024{\natexlab{a}})Zhang, Song, Wei, Yu, Lu, and Tang]{zhang2024geolrm}
Chubin Zhang, Hongliang Song, Yi Wei, Chen Yu, Jiwen Lu, and Yansong Tang.
\newblock Geolrm: Geometry-aware large reconstruction model for high-quality 3d gaussian generation.
\newblock In \emph{NeurIPS}, 2024{\natexlab{a}}.

\bibitem[Zhang et~al.(2024{\natexlab{b}})Zhang, Zou, Li, Yi, and Wang]{zhang2024transplat}
Chuanrui Zhang, Yingshuang Zou, Zhuoling Li, Minmin Yi, and Haoqian Wang.
\newblock Transplat: Generalizable 3d gaussian splatting from sparse multi-view images with transformers.
\newblock In \emph{AAAI}, 2024{\natexlab{b}}.

\bibitem[Zhang et~al.(2024{\natexlab{c}})Zhang, Bi, Tan, Xiangli, Zhao, Sunkavalli, and Xu]{zhang2024gs}
Kai Zhang, Sai Bi, Hao Tan, Yuanbo Xiangli, Nanxuan Zhao, Kalyan Sunkavalli, and Zexiang Xu.
\newblock Gs-lrm: Large reconstruction model for 3d gaussian splatting.
\newblock In \emph{ECCV}, 2024{\natexlab{c}}.

\bibitem[Zhang et~al.(2018)Zhang, Isola, Efros, Shechtman, and Wang]{zhang2018unreasonable}
Richard Zhang, Phillip Isola, Alexei~A Efros, Eli Shechtman, and Oliver Wang.
\newblock The unreasonable effectiveness of deep features as a perceptual metric.
\newblock In \emph{CVPR}, 2018.

\bibitem[Zheng et~al.(2024)Zheng, Zhou, Shao, Liu, Zhang, Nie, and Liu]{zheng2024gps}
Shunyuan Zheng, Boyao Zhou, Ruizhi Shao, Boning Liu, Shengping Zhang, Liqiang Nie, and Yebin Liu.
\newblock Gps-gaussian: Generalizable pixel-wise 3d gaussian splatting for real-time human novel view synthesis.
\newblock In \emph{CVPR}, 2024.

\bibitem[Zhou et~al.(2018)Zhou, Tucker, Flynn, Fyffe, and Snavely]{zhou2018stereo}
Tinghui Zhou, Richard Tucker, John Flynn, Graham Fyffe, and Noah Snavely.
\newblock Stereo magnification: Learning view synthesis using multiplane images.
\newblock In \emph{ACM TOG}, 2018.

\bibitem[Zou et~al.(2024)Zou, Yu, Guo, Li, Liang, Cao, and Zhang]{zou2024triplane}
Zi-Xin Zou, Zhipeng Yu, Yuan-Chen Guo, Yangguang Li, Ding Liang, Yan-Pei Cao, and Song-Hai Zhang.
\newblock Triplane meets gaussian splatting: Fast and generalizable single-view 3d reconstruction with transformers.
\newblock In \emph{CVPR}, 2024.

\end{thebibliography}
}
\clearpage
\appendix
\counterwithin{figure}{section}
\counterwithin{table}{section}
\renewcommand{\thefigure}{\thesection\arabic{figure}}
\renewcommand{\thetable}{\thesection\arabic{table}}

\section{Additional Analysis}

\subsection{Visual Encoder Analysis}

\begin{table}[!bp]
    \centering
    \resizebox{\linewidth}{!}{
        \setlength{\tabcolsep}{1pt}
        \begin{tabular}{lccc}
            \toprule
            \textbf{Encoder}                     & \textbf{Supervision} & \textbf{Dataset} & \textbf{Arch.}              \\
            \midrule
            \multicolumn{4}{l}{\color{darkred}{\textbf{\textit{Semantic-aligned}}}}                                      \\
            DeIT III~\cite{touvron2022deit}      & Classification       & ImageNet-22k     & ViT-B/16                    \\
            CLIP~\cite{radford2021learning}      & Language             & WIT-400M         & ViT-B/16                    \\
            DINOv2~\cite{oquab2023dinov2}        & Image feature        & LVD-142M         & ViT-B/14                    \\
            \midrule
            \multicolumn{4}{l}{\color{darkred}{\textbf{\textit{Spatial-aligned}}}}                                       \\
            Depth Any.~\cite{yang2024depth}      & Depth                & MIX-14           & ViT-B/14                    \\
            Depth Any. V2~\cite{yang2024depthv2} & Depth                & MIX-13           & ViT-B/14                    \\
            SAM~\cite{kirillov2023segment}       & Segmentation         & SA-1B            & ViT-B/16                    \\
            SAM 2~\cite{ravi2024sam}             & Segmentation         & SA-V             & Hiera~\cite{ryali2023hiera} \\
            DUSt3R~\cite{wang2024dust3r}         & Point regression     & MIX-8            & ViT-L/16                    \\
            MASt3R~\cite{leroy2025grounding}     & Point matching       & MIX-14           & ViT-L/16                    \\
            MAE~\cite{he2022masked}              & Pixel                & ImageNet-1k      & ViT-B/16                    \\
            SD-VAE~\cite{rombach2022high}        & Pixel                & OpenImages       & CNN                         \\
            \bottomrule
        \end{tabular}
    }
    \caption{
        \textbf{Overview of investigated visual foundation models.}
    }
    \label{tab:encoder_type}
\end{table}
\begin{table*}[!bp]
    \centering
    \resizebox{\linewidth}{!}{
        \setlength{\tabcolsep}{2pt}
        \begin{tabular}{l|ccc|ccc|ccc|ccc|ccc}
            \toprule
            \multirow{2}{*}{\textbf{Encoder}}    & \multirow{2}{*}{\textbf{Res.}} & \multirow{2}{*}{\textbf{Params.}} & \multirow{2}{*}{\textbf{Time (s)$^{\dagger}$}} & \multicolumn{3}{c|}{\textbf{RealEstate10K}} & \multicolumn{3}{c|}{\textbf{RealEstate10K $\rightarrow$ ACID}} & \multicolumn{3}{c|}{\textbf{RealEstate10K $\rightarrow$ DTU}} & \multicolumn{3}{c}{\textbf{Overall}}                                                                                                                                                                                                                                                         \\
                                                 &                                &                                   &                                                & \textbf{PSNR$\uparrow$}                     & \textbf{SSIM$\uparrow$}                                        & \textbf{LPIPS$\downarrow$}                                    & \textbf{PSNR$\uparrow$}              & \textbf{SSIM$\uparrow$}      & \textbf{LPIPS$\downarrow$}   & \textbf{PSNR$\uparrow$}      & \textbf{SSIM$\uparrow$}      & \textbf{LPIPS$\downarrow$}   & \textbf{PSNR$\uparrow$}      & \textbf{SSIM$\uparrow$}      & \textbf{LPIPS$\downarrow$}   \\
            \midrule
            \multicolumn{16}{l}{\color{darkred}{\textbf{\textit{Semantic-aligned}}}}                                                                                                                                                                                                                                                                                                                                                                                                                                                                                                                                                                 \\
            DeIT III~\cite{touvron2022deit}      & 256                            & \cellcolor{tabthird}{86 M}        & \cellcolor{tabsecond}{0.053}                   & 24.82                                       & 0.832                                                          & 0.156                                                         & 26.26                                & 0.789                        & 0.186                        & 14.44                        & 0.481                        & 0.393                        & 21.84                        & 0.701                        & 0.245                        \\
            CLIP~\cite{radford2021learning}      & 256                            & \cellcolor{tabthird}{86 M}        & \cellcolor{tabfirst}{0.051}                    & 25.14                                       & 0.839                                                          & 0.151                                                         & 26.44                                & 0.791                        & 0.184                        & 14.24                        & 0.475                        & 0.415                        & 21.94                        & 0.702                        & 0.250                        \\
            DINOv2~\cite{oquab2023dinov2}        & 224                            & \cellcolor{tabthird}{86 M}        & \cellcolor{tabfirst}{0.051}                    & 25.74                                       & 0.856                                                          & 0.140                                                         & 26.90                                & 0.809                        & 0.173                        & 14.77                        & 0.502                        & 0.368                        & 22.47                        & 0.722                        & 0.227                        \\
            \midrule
            \multicolumn{16}{l}{\color{darkred}{\textbf{\textit{Spatial-aligned}}}}                                                                                                                                                                                                                                                                                                                                                                                                                                                                                                                                                                  \\
            Depth Any.~\cite{yang2024depth}      & 224                            & \cellcolor{tabthird}{86 M}        & 0.056                                          & 26.05                                       & 0.863                                                          & 0.136                                                         & 27.20                                & 0.817                        & 0.168                        & 14.78                        & 0.491                        & 0.368                        & 22.68                        & 0.724                        & 0.224                        \\
            Depth Any. V2~\cite{yang2024depthv2} & 224                            & \cellcolor{tabthird}{86 M}        & \cellcolor{tabthird}{0.054}                    & 25.93                                       & 0.861                                                          & 0.137                                                         & 27.12                                & 0.815                        & 0.169                        & 15.07                        & 0.509                        & 0.360                        & 22.71                        & 0.728                        & 0.222                        \\
            SAM~\cite{kirillov2023segment}       & 256                            & 89 M                              & 0.065                                          & 26.39                                       & 0.869                                                          & \cellcolor{tabthird}{0.130}                                   & 27.79                                & 0.831                        & 0.158                        & 14.95                        & 0.521                        & 0.343                        & 23.04                        & 0.740                        & 0.210                        \\
            SAM 2~\cite{ravi2024sam}             & 256                            & \cellcolor{tabsecond}{68 M}       & \cellcolor{tabthird}{0.054}                    & 26.12                                       & 0.865                                                          & 0.135                                                         & 27.59                                & 0.826                        & 0.163                        & 14.88                        & 0.505                        & 0.361                        & 22.86                        & 0.732                        & 0.220                        \\
            DUSt3R~\cite{wang2024dust3r}         & 256                            & 303 M                             & 0.073                                          & \cellcolor{tabthird}{26.56}                 & \cellcolor{tabthird}{0.873}                                    & \cellcolor{tabsecond}{0.129}                                  & 27.73                                & 0.833                        & 0.158                        & \cellcolor{tabthird}{15.21}  & \cellcolor{tabthird}{0.527}  & \cellcolor{tabthird}{0.342}  & \cellcolor{tabthird}{23.17}  & 0.744                        & 0.210                        \\
            MASt3R~\cite{leroy2025grounding}     & 256                            & 303 M                             & 0.073                                          & \cellcolor{tabsecond}{26.63}                & \cellcolor{tabfirst}{0.876}                                    & \cellcolor{tabfirst}{0.127}                                   & \cellcolor{tabthird}{27.91}          & \cellcolor{tabthird}{0.837}  & \cellcolor{tabthird}{0.156}  & \cellcolor{tabfirst}{15.32}  & \cellcolor{tabsecond}{0.552} & \cellcolor{tabsecond}{0.331} & \cellcolor{tabsecond}{23.29} & \cellcolor{tabsecond}{0.755} & \cellcolor{tabsecond}{0.205} \\
            MAE~\cite{he2022masked}              & 256                            & \cellcolor{tabthird}{86 M}        & \cellcolor{tabfirst}{0.051}                    & \cellcolor{tabfirst}{26.65}                 & \cellcolor{tabsecond}{0.874}                                   & \cellcolor{tabfirst}{0.127}                                   & \cellcolor{tabsecond}{28.04}         & \cellcolor{tabsecond}{0.838} & \cellcolor{tabsecond}{0.154} & \cellcolor{tabsecond}{15.24} & 0.526                        & 0.344                        & \cellcolor{tabfirst}{23.31}  & \cellcolor{tabthird}{0.746}  & \cellcolor{tabthird}{0.208}  \\
            SD-VAE~\cite{rombach2022high}        & 256                            & \cellcolor{tabfirst}{34 M}        & \cellcolor{tabfirst}{0.051}                    & 26.50                                       & 0.872                                                          & \cellcolor{tabsecond}{0.129}                                  & \cellcolor{tabfirst}{28.16}          & \cellcolor{tabfirst}{0.842}  & \cellcolor{tabfirst}{0.153}  & \cellcolor{tabthird}{15.21}  & \cellcolor{tabfirst}{0.567}  & \cellcolor{tabfirst}{0.326}  & \cellcolor{tabsecond}{23.29} & \cellcolor{tabfirst}{0.760}  & \cellcolor{tabfirst}{0.203}  \\
            \bottomrule
        \end{tabular}
    }
    \caption{
        \textbf{Performance comparison across various visual encoders.}
        $^{\dagger}$: Inference time of the entire network.
    }
    \label{tab:encoder_performance}
\end{table*}
\begin{table*}[!htbp]
    \centering
    \begin{NiceTabular}{c|ccc|ccc|ccc}
        \CodeBefore
        \rowcolor{heavygray}{3-3}
        \Body

        \toprule
        \Block{2-1}{\textbf{Finetuning}} & \Block{1-3}{\textbf{RealEstate10K}} &                         &                            & \Block{1-3}{\textbf{RealEstate10K $\to$ ACID}} &                         &                            & \Block{1-3}{\textbf{RealEstate10K $\to$ DTU}}                                                        \\
                                         & \textbf{PSNR$\uparrow$}             & \textbf{SSIM$\uparrow$} & \textbf{LPIPS$\downarrow$} & \textbf{PSNR$\uparrow$}                        & \textbf{SSIM$\uparrow$} & \textbf{LPIPS$\downarrow$} & \textbf{PSNR$\uparrow$}                       & \textbf{SSIM$\uparrow$} & \textbf{LPIPS$\downarrow$} \\
        \midrule
        \xmark                           & 26.68                               & 0.875                   & 0.126                      & 28.29                                          & 0.844                   & 0.152                      & 15.38                                         & 0.587                   & 0.317                      \\
        \midrule
        \Block{2-1}{\cmark}              & \textbf{26.81}                      & \textbf{0.878}          & \textbf{0.125}             & \textbf{28.39}                                 & \textbf{0.846}          & \textbf{0.151}             & \textbf{15.63}                                & \textbf{0.598}          & \textbf{0.309}             \\[-3pt]
                                         & \diff{+0.13}                        & \diff{+0.003}           & \diff{-0.001}              & \diff{+0.10}                                   & \diff{+0.002}           & \diff{-0.001}              & \diff{+0.25}                                  & \diff{+0.011}           & \diff{-0.008}              \\
        \bottomrule
    \end{NiceTabular}
    \caption{
        \textbf{Effect of finetuning the SD-VAE encoder.}
    }
    \label{tab:finetuning}
\end{table*}

\noindent\textbf{Visual encoder comparison.}
As summarized in~\cref{tab:encoder_type}, we benchmark a diverse collection of visual foundation models spanning two supervision paradigms.
\textbf{Semantic-aligned} encoders, such as DeiT III~\cite{touvron2022deit}, CLIP~\cite{radford2021learning}, and DINOv2~\cite{oquab2023dinov2}, are trained on image-level objectives to learn high-level semantic representations.
For instance, CLIP aligns vision and language through contrastive learning, while DINOv2 learns robust self-supervised features from massive unlabeled datasets.
In contrast, \textbf{spatial-aligned} models leverage pixel-level supervision to capture fine-grained geometry and detail.
This category includes models for monocular depth estimation (Depth Anything~\cite{yang2024depth, yang2024depthv2}), semantic segmentation (SAM~\cite{kirillov2023segment, ravi2024sam}), dense 3D reconstruction (DUSt3R~\cite{wang2024dust3r} and MASt3R~\cite{leroy2025grounding}), and self-supervised image reconstruction (MAE~\cite{he2022masked}).
SD-VAE~\cite{rombach2022high}, the compact and efficient variational autoencoder from the Stable Diffusion~\cite{rombach2022high} pipeline, also falls into this category.

As shown in~\cref{fig:encoder_performance} of the main paper and detailed in~\cref{tab:encoder_performance}, our evaluation reveals key insights into foundation model effectiveness for 3D reconstruction.
Spatial-aligned encoders consistently outperform semantic-aligned counterparts across all metrics, achieving higher PSNR (22.68-23.39 vs.
21.84-22.47 ) and lower LPIPS (0.203-0.224 vs. 0.227-0.250).
This performance gap indicates that pixel-level supervision provides richer geometric priors than semantic-level training for 3D reconstruction.
We also observe that model size does not correlate with reconstruction quality.
For instance, while large models like MASt3R~\cite{leroy2025grounding} (303M) achieve a top-tier PSNR of 23.29 , they do so at significant computational cost (73ms).
In contrast, SD-VAE achieves comparable PSNR while delivering superior SSIM and LPIPS scores with a model nearly 9$\times$ smaller (34M) and 30\% faster inference (51ms).
\textbf{SD-VAE emerges as the Pareto-optimal choice, delivering the best reconstruction quality and computational efficiency among all evaluated encoders.}
Cross-domain evaluation demonstrates SD-VAE's strong generalization: it not only excels on RealEstate10K but also maintains robust performance on challenging out-of-distribution datasets including outdoor ACID scenes and object-centric DTU views.
Based on its superior accuracy, parameter efficiency, and cross-domain transfer, we adopt SD-VAE as our default visual encoder.

\noindent\textbf{Encoder Adaptation Analysis.}
To assess the potential for further performance gains, we evaluate the effect of finetuning the SD-VAE encoder, with results detailed in~\cref{tab:finetuning}.
Finetuning yields consistent improvements across all scenarios, with the gains being most pronounced in cross-domain generalization.
On the challenging DTU dataset, for instance, performance improves across all metrics (+0.25 PSNR, +0.011 SSIM, -0.008 LPIPS).
These results demonstrate that encoder adaptation can enhance reconstruction quality, particularly for out-of-distribution scenarios.
However, we choose to keep the encoder frozen to preserve pretrained representations and avoid dataset-specific overfitting.
While this prioritizes generalization over peak performance, encoder adaptation remains a promising avenue for future work.

\subsection{Training Dynamics}

\begin{figure*}[!htbp]
    \centering
    \begin{subfigure}
        {0.33\textwidth}
        \centering
        \includegraphics[width=\linewidth]{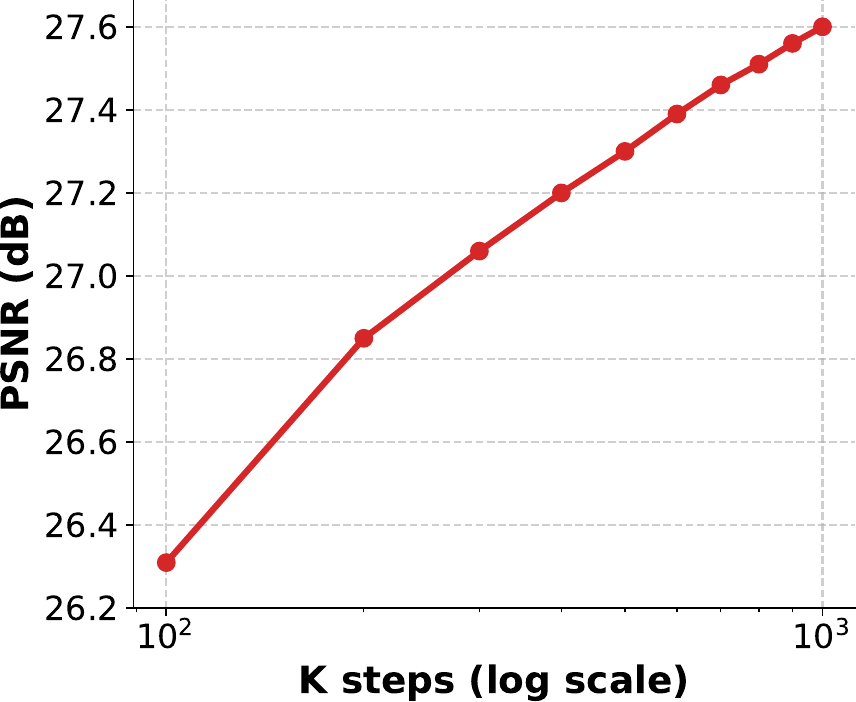}
    \end{subfigure}
    \begin{subfigure}
        {0.33\textwidth}
        \centering
        \includegraphics[width=\linewidth]{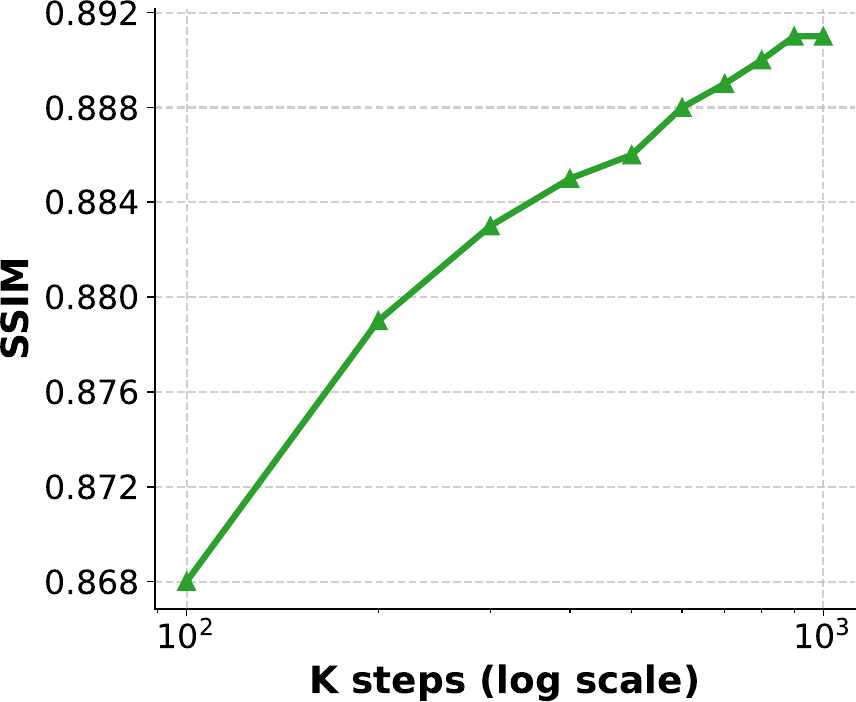}
    \end{subfigure}
    \begin{subfigure}
        {0.33\textwidth}
        \centering
        \includegraphics[width=\linewidth]{
            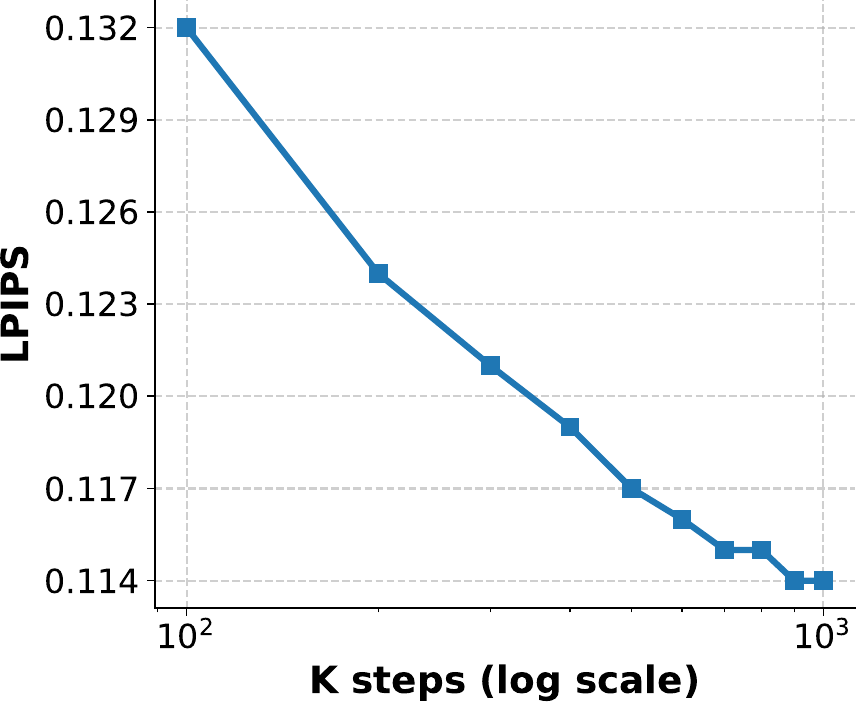
        }
    \end{subfigure}
    \caption{
        \textbf{Performance scaling during pre-training.}
        H3R exhibits power-law scaling on RealEstate10K, where performance improves linearly with the logarithm of training steps, similar to scaling patterns observed in large language models~\cite{kaplan2020scaling,touvron2023llama,hoffmann2022training}.
        PSNR shows no signs of saturation even at 1 million training steps, indicating potential for further improvement.
    }
    \label{fig:performance_scaling}
\end{figure*}

\noindent\textbf{Performance scaling during pre-training.}
We analyze the scaling behavior of our model during pre-training on RealEstate10K, as shown in~\cref{fig:performance_scaling}.
The results reveal a predictable power-law scaling relationship, where performance across all three metrics improves near-linearly with the logarithm of training steps.
This scaling behavior follows the well-documented power-law relationship observed in large language models~\cite{kaplan2020scaling,hoffmann2022training}, suggesting that 3D reconstruction models benefit from similar scaling principles.
Crucially, even after 1 million training steps, the PSNR curve does not yet show signs of saturation, indicating potential for further improvement.
The steady, predictable scaling suggests that our 3D reconstruction model benefits from the same fundamental scaling laws that govern other foundation models, highlighting the value of increased computational investment.

\noindent\textbf{Effect of Exponential Moving Average (EMA).}
We evaluate the impact of applying EMA to model parameters during training, with results presented in~\cref{tab:ema}.
EMA stabilizes training by maintaining exponentially weighted averages of model parameters, effectively reducing parameter noise and promoting convergence to more generalizable solutions.
This technique has proven effective across various vision tasks, particularly in image generation~\cite{rombach2022high} and restoration~\cite{wang2021real}.
Our results demonstrate consistent improvements across all metrics: PSNR increases by 0.18 , SSIM by 0.002, and LPIPS decreases by 0.002.
Given these consistent improvements, we adopt EMA by default in our final model.

\begin{table}[!htbp]
    \centering
    \begin{NiceTabular}{c|ccc}
        \CodeBefore
        \rowcolor{heavygray}{3-4}
        \Body

        \toprule
        \textbf{EMA}        & \textbf{PSNR$\uparrow$} & \textbf{SSIM$\uparrow$} & \textbf{LPIPS$\downarrow$} \\
        \midrule
        \xmark              & 27.42                   & 0.889                   & 0.116                      \\
        \midrule
        \Block{2-1}{\cmark} & \textbf{27.60}          & \textbf{0.891}          & \textbf{0.114}             \\[-3pt]
                            & \diff{+0.18}            & \diff{+0.002}           & \diff{-0.002}              \\
        \bottomrule
    \end{NiceTabular}
    \caption{
        \textbf{Effect of EMA on RealEstate10K.}
    }
    \label{tab:ema}
\end{table}

\subsection{Input Analysis}

\noindent\textbf{Effect of camera pose normalization.}
We analyze the effect of camera pose normalization, a common technique in 3D reconstruction~\cite{mildenhall2019local,zhang2024gs} where input poses are transformed to a canonical coordinate frame defined by their mean pose.
As shown in~\cref{tab:camera_norm}, normalization exhibits contrasting effects across datasets.
While normalization yields marginal improvements on large-scale, forward-facing scenes like RealEstate10K and ACID, it significantly degrades performance on the object-centric DTU dataset, causing a -0.39 drop in PSNR.
We attribute this discrepancy stems from different camera motion characteristics.
In large scenes, normalization stabilizes training by bounding the coordinate space.
For object-centric settings, however, poses are already tightly clustered around the object, and normalization can amplify small but meaningful positional variations, disrupting the geometric cues essential for precise reconstruction.
Given our emphasis on cross-domain generalization, we omit pose normalization in our final model.

\begin{table*}[!htbp]
    \centering
    \begin{NiceTabular}{c|ccc|ccc|ccc}
        \CodeBefore
        \rowcolor{heavygray}{3-3}
        \Body

        \toprule
        \textbf{Camera pose}   & \Block{1-3}{\textbf{RealEstate10K}} &                         &                            & \Block{1-3}{\textbf{RealEstate10K $\to$ ACID}} &                         &                            & \Block{1-3}{\textbf{RealEstate10K $\to$ DTU}}                                                        \\
        \textbf{normalization} & \textbf{PSNR$\uparrow$}             & \textbf{SSIM$\uparrow$} & \textbf{LPIPS$\downarrow$} & \textbf{PSNR$\uparrow$}                        & \textbf{SSIM$\uparrow$} & \textbf{LPIPS$\downarrow$} & \textbf{PSNR$\uparrow$}                       & \textbf{SSIM$\uparrow$} & \textbf{LPIPS$\downarrow$} \\
        \midrule
        \xmark                 & 27.05                               & 0.882                   & 0.121                      & 28.55                                          & 0.849                   & 0.145                      & \textbf{15.49}                                & \textbf{0.587}          & \textbf{0.314}             \\
        \midrule
        \Block{2-1}{\cmark}    & \textbf{27.08}                      & \textbf{0.883}          & \textbf{0.120}             & \textbf{28.61}                                 & \textbf{0.852}          & \textbf{0.144}             & 15.10                                         & 0.565                   & 0.319                      \\[-3pt]
                               & \diff{+0.03}                        & \diff{+0.001}           & \diff{-0.001}              & \diff{+0.06}                                   & \diff{+0.003}           & \diff{-0.001}              & \diff{-0.39}                                  & \diff{-0.022}           & \diff{+0.005}              \\
        \bottomrule
    \end{NiceTabular}
    \caption{
        \textbf{Effect of camera pose normalization.}
        While camera normalization yields modest improvements on scene-level datasets, it significantly degrades generalization performance on the object-level DTU dataset.
    }
    \label{tab:camera_norm}
\end{table*}

\noindent\textbf{Effect of input resolution.}
We investigate the impact of input resolution on reconstruction performance by comparing models trained at 224$\times$224 versus 256$\times$256 resolution, as shown in~\cref{tab:resolution}.
Higher resolution yields only modest improvements: +0.17 PSNR for MASt3R, +0.12 PSNR for DUSt3R, and +0.24 PSNR for CLIP on RealEstate10K.
The results suggest that input resolution is less critical for reconstruction quality compared to architectural choice and training method.
We adopt 256$\times$256 as our default resolution primarily for consistency with standard practice~\cite{charatan2024pixelsplat,chen2024mvsplat}.

\begin{table*}[!htbp]
    \centering
    \resizebox{\linewidth}{!}{
        \begin{tabular}{l|c|ccc|ccc|ccc}
            \toprule
            \multirow{2}{*}{\textbf{Encoder}}                 & \multirow{2}{*}{\textbf{Res.}} & \multicolumn{3}{c|}{\textbf{RealEstate10K}} & \multicolumn{3}{c|}{\textbf{RealEstate10K $\to$ ACID}} & \multicolumn{3}{c}{\textbf{RealEstate10K $\to$ DTU}}                                                                                                                                                                                     \\
                                                              &                                & \textbf{PSNR$\uparrow$}                     & \textbf{SSIM$\uparrow$}                                & \textbf{LPIPS$\downarrow$}                           & \textbf{PSNR$\uparrow$}     & \textbf{SSIM$\uparrow$}     & \textbf{LPIPS$\downarrow$}  & \textbf{PSNR$\uparrow$}     & \textbf{SSIM$\uparrow$}     & \textbf{LPIPS$\downarrow$}  \\
            \midrule
            DINOv2~\cite{oquab2023dinov2}                     & 224                            & 25.74                                       & 0.856                                                  & 0.140                                                & 26.90                       & 0.809                       & 0.173                       & 14.77                       & 0.502                       & 0.368                       \\
            Depth Any.~\cite{yang2024depth}                   & 224                            & 26.05                                       & 0.863                                                  & 0.136                                                & 27.20                       & 0.817                       & 0.168                       & 14.78                       & 0.491                       & 0.368                       \\
            Depth Any. V2~\cite{yang2024depthv2}              & 224                            & 25.93                                       & 0.861                                                  & 0.137                                                & 27.12                       & 0.815                       & 0.169                       & 15.07                       & 0.509                       & 0.360                       \\
            \midrule
            \multirow{2}{*}{CLIP~\cite{radford2021learning}}  & 224                            & 24.90                                       & 0.832                                                  & 0.155                                                & 26.20                       & 0.782                       & 0.189                       & 13.91                       & 0.448                       & 0.437                       \\
                                                              & 256                            & \cellcolor{tabfirst}{25.14}                 & \cellcolor{tabfirst}{0.839}                            & \cellcolor{tabfirst}{0.151}                          & \cellcolor{tabfirst}{26.44} & \cellcolor{tabfirst}{0.791} & \cellcolor{tabfirst}{0.184} & \cellcolor{tabfirst}{14.24} & \cellcolor{tabfirst}{0.475} & \cellcolor{tabfirst}{0.415} \\
            \midrule
            \multirow{2}{*}{DUSt3R~\cite{wang2024dust3r}}     & 224                            & 26.44                                       & 0.871                                                  & 0.130                                                & 27.63                       & 0.830                       & 0.160                       & \cellcolor{tabfirst}{15.24} & \cellcolor{tabfirst}{0.530} & 0.346                       \\
                                                              & 256                            & \cellcolor{tabfirst}{26.56}                 & \cellcolor{tabfirst}{0.873}                            & \cellcolor{tabfirst}{0.129}                          & \cellcolor{tabfirst}{27.73} & \cellcolor{tabfirst}{0.833} & \cellcolor{tabfirst}{0.158} & 15.21                       & 0.527                       & \cellcolor{tabfirst}{0.342} \\
            \midrule
            \multirow{2}{*}{MASt3R~\cite{leroy2025grounding}} & 224                            & 26.46                                       & 0.872                                                  & 0.130                                                & 27.79                       & 0.834                       & 0.158                       & \cellcolor{tabfirst}{15.32} & 0.541                       & 0.335                       \\
                                                              & 256                            & \cellcolor{tabfirst}{26.63}                 & \cellcolor{tabfirst}{0.876}                            & \cellcolor{tabfirst}{0.127}                          & \cellcolor{tabfirst}{27.91} & \cellcolor{tabfirst}{0.837} & \cellcolor{tabfirst}{0.156} & \cellcolor{tabfirst}{15.32} & \cellcolor{tabfirst}{0.552} & \cellcolor{tabfirst}{0.331} \\
            \bottomrule
        \end{tabular}
    }
    \caption{
        \textbf{Effect of input resolution.}
    }
    \label{tab:resolution}
\end{table*}
\begin{table*}[!htbp]
    \centering
    \resizebox{\textwidth}{!}{
        \begin{NiceTabular}{c|ccc|ccc|ccc|ccc}
            \CodeBefore
            \rowcolor{heavygray}{4-5}
            \Body

            \toprule
            \Block{2-1}{\textbf{Method}}     & \multicolumn{3}{c|}{\textbf{RealEstate10K (2 views)}} & \multicolumn{3}{c|}{\textbf{RealEstate10K (4 views)}} & \multicolumn{3}{c|}{\textbf{RealEstate10K (6 views)}} & \multicolumn{3}{c}{\textbf{RealEstate10K (8 views)}}                                                                                                                                                                                                                          \\
                                             & \textbf{PSNR$\uparrow$}                               & \textbf{SSIM$\uparrow$}                               & \textbf{LPIPS$\downarrow$}                            & \textbf{PSNR$\uparrow$}                              & \textbf{SSIM$\uparrow$} & \textbf{LPIPS$\downarrow$} & \textbf{PSNR$\uparrow$} & \textbf{SSIM$\uparrow$} & \textbf{LPIPS$\downarrow$} & \textbf{PSNR$\uparrow$} & \textbf{SSIM$\uparrow$} & \textbf{LPIPS$\downarrow$} \\
            \midrule
            MVSplat                          & 26.36                                                 & 0.868                                                 & 0.129                                                 & 22.20                                                & 0.820                   & 0.118                      & 20.95                   & 0.803                   & 0.203                      & 20.35                   & 0.792                   & 0.214                      \\
            \midrule
            \Block{2-1}{H3R-$\alpha$ (Ours)} & \textbf{27.46}                                        & \textbf{0.889}                                        & \textbf{0.115}                                        & \textbf{29.28}                                       & \textbf{0.920}          & \textbf{0.090}             & \textbf{29.97}          & \textbf{0.930}          & \textbf{0.083}             & \textbf{30.24}          & \textbf{0.934}          & \textbf{0.080}             \\[-3pt]
                                             & \diff{+1.10}                                          & \diff{+0.021}                                         & \diff{-0.014}                                         & \diff{+7.08}                                         & \diff{+0.100}           & \diff{-0.028}              & \diff{+9.02}            & \diff{+0.127}           & \diff{-0.120}              & \diff{+9.89}            & \diff{+0.142}           & \diff{-0.134}              \\
            \bottomrule
        \end{NiceTabular}
    }
    \caption{
        \textbf{Performance comparison across varying number of input views.}
    }
    \label{tab:view_scaling}
\end{table*}

\noindent\textbf{Effect of number of input views.}
We evaluate our model's adaptability with respect to the number of input views, comparing H3R-$\alpha$ against MVSplat on RealEstate10K.
As detailed in~\cref{tab:view_scaling}, the two methods exhibit opposite scaling behaviors.
H3R-$\alpha$ demonstrates robustness with the number of input views, with PSNR climbing from 27.46 (2 views) to 30.24 (8 views).
Conversely, MVSplat degrades as more views are added, dropping from 22.20 (4 views) to 20.35 (8 views).
These results highlight our framework's effective multi-view aggregation capabilities.

\noindent\textbf{Effect of view overlap.}
We evaluate our method's robustness to varying view overlap, comparing H3R against MVSplat in~\cref{tab:overlap}.
H3R consistently outperforms the baseline across all tested overlap ranges.
The performance gain is most pronounced under challenging low-overlap scenarios [0.60,0.65], where our method achieves a substantial 1.73 dB improvement in PSNR.
As the view overlap increases, providing richer geometric cues, this advantage gradually narrows to 1.07 in high-overlap conditions [0.95,1.00].
This trend demonstrates H3R's robustness across varying geometric constraints, with particularly strong performance in challenging scenarios where existing methods often fail.

\begin{table*}[!htbp]
    \centering
    \begin{tabular}{c|ccc|ccc|ccc}
        \toprule
        \multirow{2}{*}{\textbf{Overlap}} & \multicolumn{3}{c|}{\textbf{MVSplat}} & \multicolumn{3}{c|}{\textbf{H3R (ours)}} & \multicolumn{3}{c}{\textbf{Improvement}}                                                                                                                                                                                           \\
                                          & \textbf{PSNR$\uparrow$}               & \textbf{SSIM$\uparrow$}                  & \textbf{LPIPS$\downarrow$}               & \textbf{PSNR$\uparrow$} & \textbf{SSIM$\uparrow$} & \textbf{LPIPS$\downarrow$} & \textbf{$\Delta$PSNR$\uparrow$} & \textbf{$\Delta$SSIM$\uparrow$} & \textbf{$\Delta$LPIPS$\downarrow$} \\
        \midrule
        $[0.60, 0.65)$                    & 24.41                                 & 0.841                                    & 0.151                                    & 26.14                   & 0.877                   & 0.127                      & +1.73                           & +0.036                          & -0.024                             \\
        $[0.65, 0.70)$                    & 25.00                                 & 0.852                                    & 0.144                                    & 26.44                   & 0.882                   & 0.124                      & +1.44                           & +0.030                          & -0.020                             \\
        $[0.70, 0.75)$                    & 25.68                                 & 0.862                                    & 0.133                                    & 27.00                   & 0.887                   & 0.116                      & +1.32                           & +0.025                          & -0.017                             \\
        $[0.75, 0.80)$                    & 25.74                                 & 0.864                                    & 0.131                                    & 27.01                   & 0.888                   & 0.115                      & +1.27                           & +0.024                          & -0.016                             \\
        $[0.80, 0.85)$                    & 26.15                                 & 0.871                                    & 0.129                                    & 27.33                   & 0.893                   & 0.113                      & +1.18                           & +0.022                          & -0.016                             \\
        $[0.85, 0.90)$                    & 26.18                                 & 0.872                                    & 0.129                                    & 27.39                   & 0.894                   & 0.114                      & +1.21                           & +0.022                          & -0.015                             \\
        $[0.90, 0.95)$                    & 25.92                                 & 0.863                                    & 0.134                                    & 27.12                   & 0.886                   & 0.118                      & +1.20                           & +0.023                          & -0.016                             \\
        $[0.95, 1.00)$                    & 27.86                                 & 0.881                                    & 0.117                                    & 28.93                   & 0.899                   & 0.105                      & +1.07                           & +0.018                          & -0.012                             \\
        \bottomrule
    \end{tabular}
    \caption{
        \textbf{Performance comparison across varying overlaps.}
    }
    \label{tab:overlap}
\end{table*}

\subsection{Further Comparison}

\noindent\textbf{Comparison with GS-LRM.}
We compare our H3R-$\beta$ against the state-of-the-art GS-LRM~\cite{zhang2024gs}, as shown in~\cref{tab:gslrm}.
Our method demonstrates remarkable efficiency, utilizing only 30\% of the trainable parameters (91M vs.
300M) and 20\% of the training cost (37 vs. 192 GPU-days).
This efficiency does not compromise quality; in fact, our method achieves superior perceptual quality with better SSIM (0.897 vs.
0.892) and LPIPS (0.110 vs. 0.114) scores.
This combination of efficiency and quality makes our model more practical for widespread adoption.

\begin{table}[!htbp]
    \centering
    \resizebox{\linewidth}{!}{
        \setlength{\tabcolsep}{3pt}
        \begin{tabular}{c|cc|ccc}
            \toprule
            \multirow{2}{*}{\textbf{Method}} & \textbf{\#Trainable} & \textbf{\#GPU days}  & \multicolumn{3}{c}{\textbf{RealEstate10K}}                                                        \\
                                             & \textbf{Params.}     & \textbf{4090~/~A100} & \textbf{PSNR$\uparrow$}                    & \textbf{SSIM$\uparrow$} & \textbf{LPIPS$\downarrow$} \\
            \midrule
            GS-LRM                           & 300M                 & 0~/~192              & \textbf{28.10}                             & 0.892                   & 0.114                      \\
            H3R-$\beta$                      & 91M                  & 30~/~7               & 28.03                                      & \textbf{0.897}          & \textbf{0.110}             \\
            \bottomrule
        \end{tabular}
    }
    \caption{
        \textbf{Comparison with GS-LRM~\cite{zhang2024gs}.}
        We achieve comparable performance with \textbf{30\%} trainable parameters and \textbf{20\%} training cost.
    }
    \label{tab:gslrm}
\end{table}

\section{3D Gaussian Parameterization}
\label{sec:gaussian_parameterization}

3D Gaussians provide an explicit and flexible representation for 3D scenes, and their parameterization is crucial for model performance.
To ensure reproducibility, we provide detailed specifications for each Gaussian parameter.
The specific configurations for each parameter are provided in~\cref{tab:gaussian}.

\begin{table}[!htbp]
    \centering
    \begin{tabular}{c|cc}
        \toprule
        \textbf{Parameter} & \textbf{Activation} & \textbf{Channel} \\
        \midrule
        Center             & None                & 3                \\
        Scale              & Sigmoid             & 3                \\
        Rotation           & L2 norm             & 4                \\
        Opacity            & Sigmoid             & 1                \\
        Color              & ReLU                & 3                \\
        \bottomrule
    \end{tabular}
    \caption{
        \textbf{3D Gaussian parameterization.}
    }
    \label{tab:gaussian}
\end{table}

\noindent\textbf{Ray distance.}
We uniformly sample 128 depth hypotheses $\{d_{i}\}_{i=1}^{128}$ in inverse depth space between the near and far planes.
The model output is transformed into a probability distribution $\omega$ over these hypotheses using softmax activation and the ray distance $t$ is computed as the weighted sum:
\begin{align}
    \omega & = \text{softmax}(\mathbf{G}_{\text{distance}}), \\
    t      & = \sum_{i=1}^{128}\omega_{i} \cdot d_{i},
\end{align}
where $\mathbf{G}_{\text{distance}}$ is the depth head output.
The near and far planes are dataset-specific.
For scene-level datasets such as RealEstate10K and ACID, we set the planes to 1 and 100, respectively.
For the object-level DTU dataset, we adopt the configuration from MVSplat~\cite{chen2024mvsplat} with near and far planes of 2.215 and 4.525, respectively.
Our pilot experiments demonstrate that employing multiple hypotheses yields slight performance improvements over the two-hypothesis approach used in GS-LRM~\cite{zhang2024gs}.

\noindent\textbf{Scale.}
Following pixelSplat~\cite{charatan2024pixelsplat}, we parameterize Gaussian scales in image space instead of world space.
The scale head output $\mathbf{G}_{\text{scale}}$ is mapped to a predefined scale range in pixel space $[s_{\text{min}}, s_{\text{max}}]$ using sigmoid activation:
\begin{align}
    \omega           & = \sigma(\mathbf{G}_{\text{scale}}),                   \\
    s_{\text{pixel}} & = (1 - \omega) s_{\text{min}} + \omega s_{\text{max}}.
\end{align}
We then compute the world-space scale $s_{\text{world}}$ as:
\begin{equation}
    s_{\text{world}}= s_{\text{pixel}} \cdot p_{\text{world}} \cdot t,
\end{equation}
where $p_{\text{world}}$ is the pixel size in world space.
This scaling approach maintains proper perspective by ensuring that distant Gaussians have appropriate screen-space sizes.
For both scene-level and object-level datasets, we set the pixel-space scale range to $s_{\text{min}} = 0.5$ and $s_{\text{max}} = 15.0$.

\noindent\textbf{Opacity.}
The opacity of each Gaussian is transformed to the range $(0, 1)$ using sigmoid activation.

\noindent\textbf{Rotation.}
As in~\cite{zhang2024gs}, we predict unnormalized quaternions and apply L2-normalization to obtain unit quaternions.

\noindent\textbf{RGB.}
For simplicity, we predict the zero-order Spherical Harmonics (SH) coefficients.
We apply ReLU activation to ensure non-negative color values.

\noindent\textbf{Center.}
Rather than predicting the Gaussian center directly, we derive it from the ray distance and camera parameters.
For each pixel, the ray origin $\text{ray}_{o}$ and direction $\text{ray}_{d}$ are computed from the known camera parameters.
The Gaussian center $xyz$ is then determined by:
\begin{equation}
    xyz = \text{ray}_{o} + t \cdot \text{ray}_{d}.
\end{equation}

\section{Implementation Details}
\label{sec:implementation_details}

\subsection{Datasets}
We train and evaluate our method on two large-scale datasets: RealEstate10K~\cite{zhou2018stereo} and ACID~\cite{liu2021infinite}.
RealEstate10K contains 67,477 training scenes and 7,289 test scenes of diverse indoor and outdoor environments from YouTube, while ACID comprises 11,075 training scenes and 1,972 test scenes of natural landscapes captured by drones.
For both datasets, camera poses are estimated using Structure-from-Motion (SfM)~\cite{schoenberger2016sfm}.
We follow the official train/test splits and evaluation protocol of pixelSplat~\cite{charatan2024pixelsplat}, where two input context views are used to synthesize three novel views for each test scene.
To evaluate cross-dataset generalization, we perform zero-shot evaluation on the object-centric DTU dataset~\cite{jensen2014large}.
Following the setup in~\cite{chen2024mvsplat}, we evaluate on 16 validation scenes, rendering four novel views for each scene.
We evaluate rendering quality with three standard metrics: PSNR, SSIM~\cite{wang2004image}, and LPIPS~\cite{zhang2018unreasonable}.

\subsection{Model Details}
Our camera-aware Transformer comprises 12 layers with hidden dimensions of 512 and MLP hidden dimensions of 1536, employing Pre-LayerNorm, QK-Norm~\cite{henry2020query}, and SwiGLU~\cite{shazeer2020glu} activation.

\subsection{Training Details}
We initialize the visual encoder from publicly available checkpoints and freeze its parameters throughout training.
Unless otherwise specified, we adopt the hyperparameters from MVSplat~\cite{chen2024mvsplat}.
Following~\cite{charatan2024pixelsplat,chen2024mvsplat}, we apply random horizontal flipping for data augmentation.
The pixel gradient loss weight is empirically set to 1.0.
We employ Bfloat16 mixed-precision training and cache visual features to accelerate training.
Detailed training settings for the RealEstate10K and ACID datasets are provided in~\cref{tab:training_setting_re10k} and~\cref{tab:training_setting_acid}, respectively.

\noindent\textbf{H3R: Pre-training (256$\times$256, 2 views)}
We pre-train the H3R model with two context views at 256$\times$256 resolution.
The model is trained for 1M steps on RealEstate10K and 400K steps on ACID.
Training requires seven days and three days, respectively, on 4 NVIDIA RTX 4090 GPUs.
Following pixelSplat~\cite{charatan2024pixelsplat}, the maximum frame distance is linearly increased from 25 to 45 over the initial 150K steps and then held constant.

\noindent\textbf{H3R-$\alpha$: Multi-view (256$\times$256, 2-8 views)}
We finetune base model with random 2-8 context views at 256$\times$256 resolution.
The model is trained for 30K steps on RealEstate10K and 90K steps on ACID.
Training takes about 15 hours on 4 NVIDIA A6000 GPUs and 28 hours on 8 NVIDIA RTX4090 GPUs.
During finetuning, we randomly include target camera poses as input with probability 0.5 for each training sample.

\noindent\textbf{H3R-$\beta$: High-resolution (512$\times$512, 2 views)}
We finetune base model with two context views at 512$\times$512 resolution.
The model is trained for 80K steps on RealEstate10K and 20K steps on ACID.
Training takes about 42 and 11 hours on 4 NVIDIA A100 GPUs, respectively.
The maximum frame distance between context views is fixed at 45.

\section{Additional Visualizations}

\begin{figure*}[!htbp]
    \centering
    \includegraphics[width=1.0\linewidth]{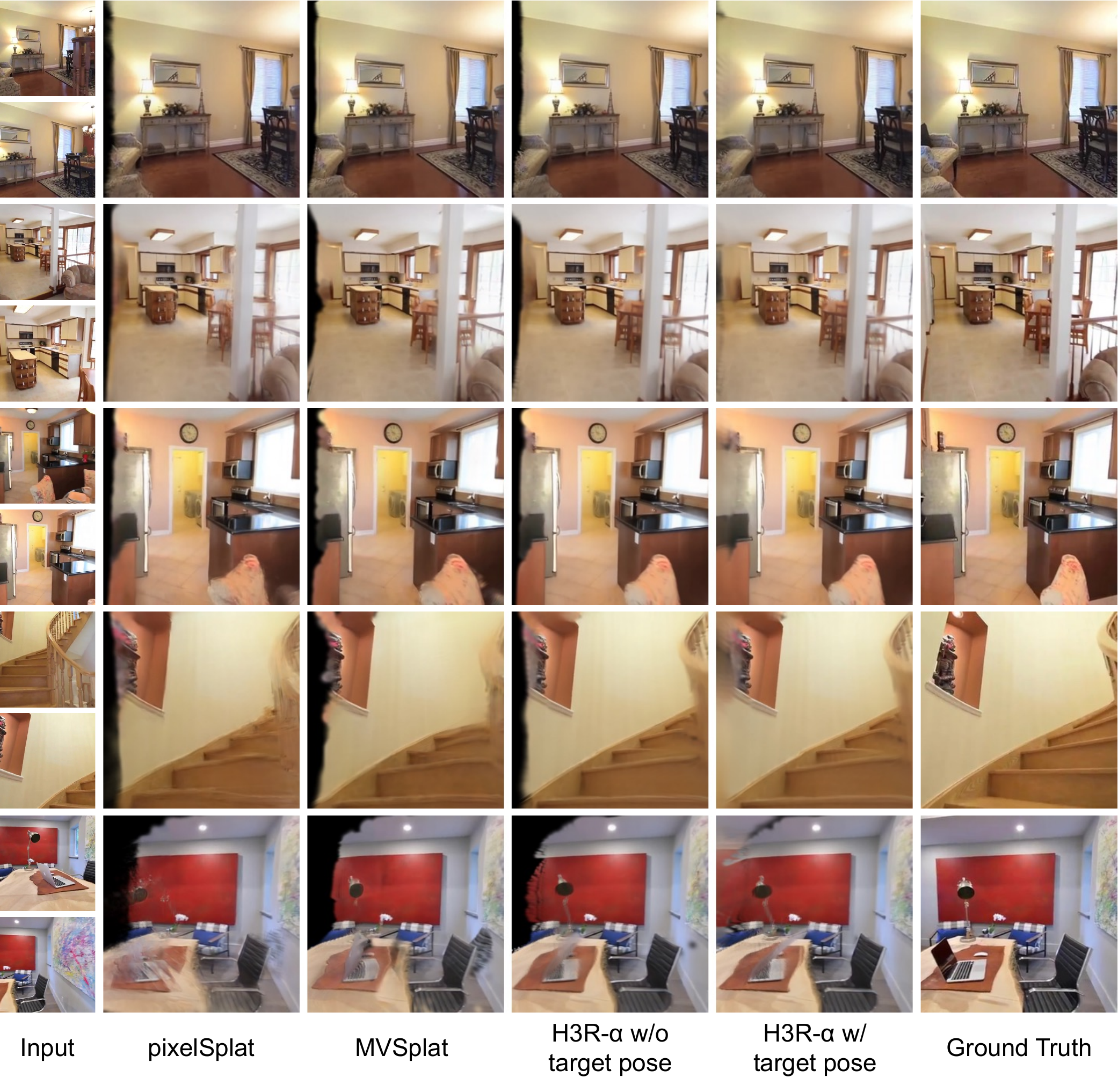}
    \caption{
        \textbf{Impact of target camera poses on RealEstate10K.}
        Our method (H3R-$\alpha$) leverages target camera poses to generate more complete and view-aligned Gaussian splats, improving geometric coherence while mitigating artifacts, particularly in unobserved regions.
    }
    \label{fig:target_pose}
\end{figure*}

\begin{figure*}[!htbp]
    \centering
    \includegraphics[width=1.0\linewidth]{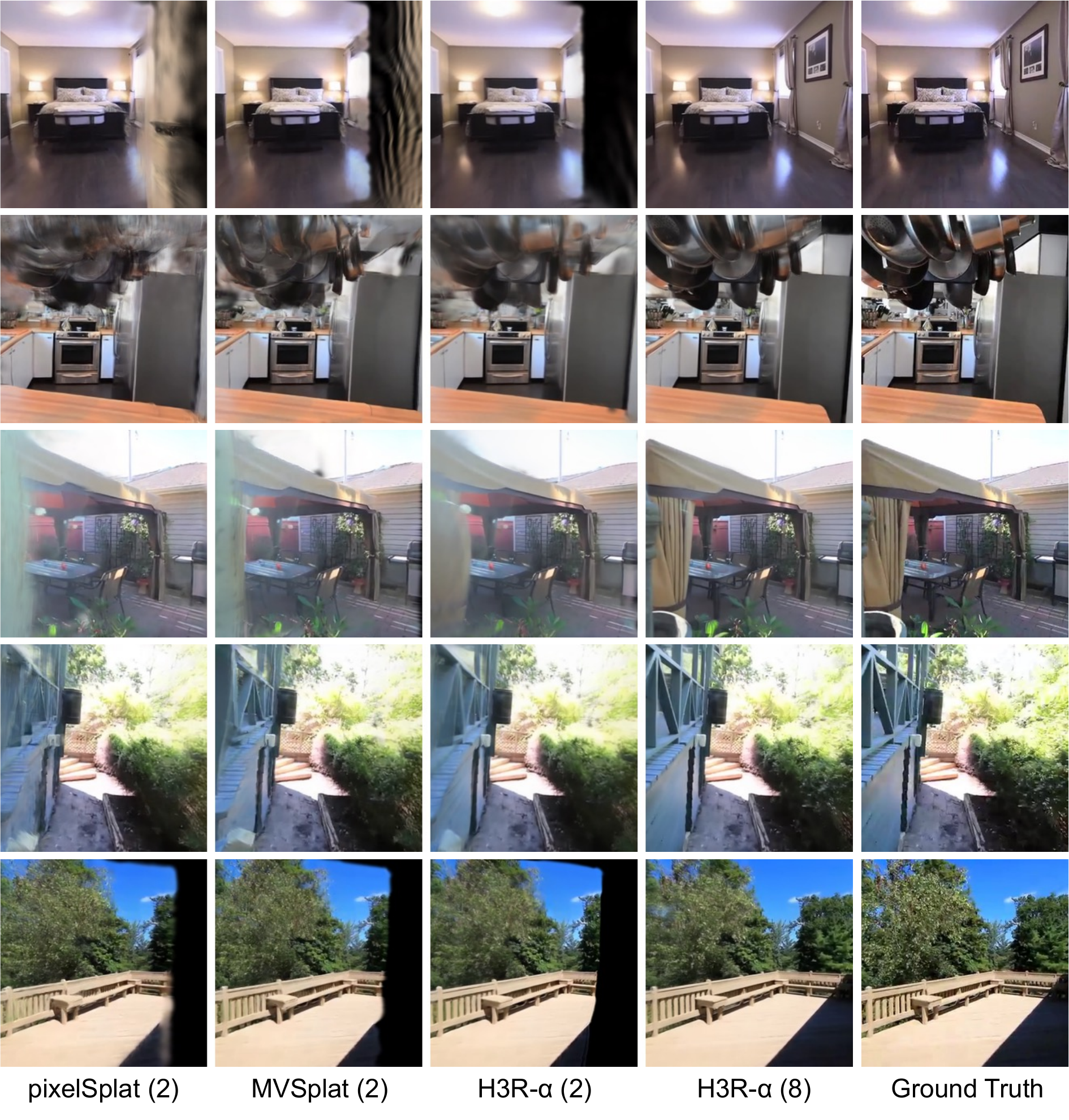}
    \caption{
        \textbf{Impact of the number of input views on RealEstate10K.}
        Increasing input views from two to eight enhances geometric completeness and visual fidelity, particularly for scene boundaries and specular surfaces.
    }
    \label{fig:view_scaling_visual}
\end{figure*}

\begin{figure*}[!htbp]
    \centering
    \includegraphics[width=1.0\linewidth]{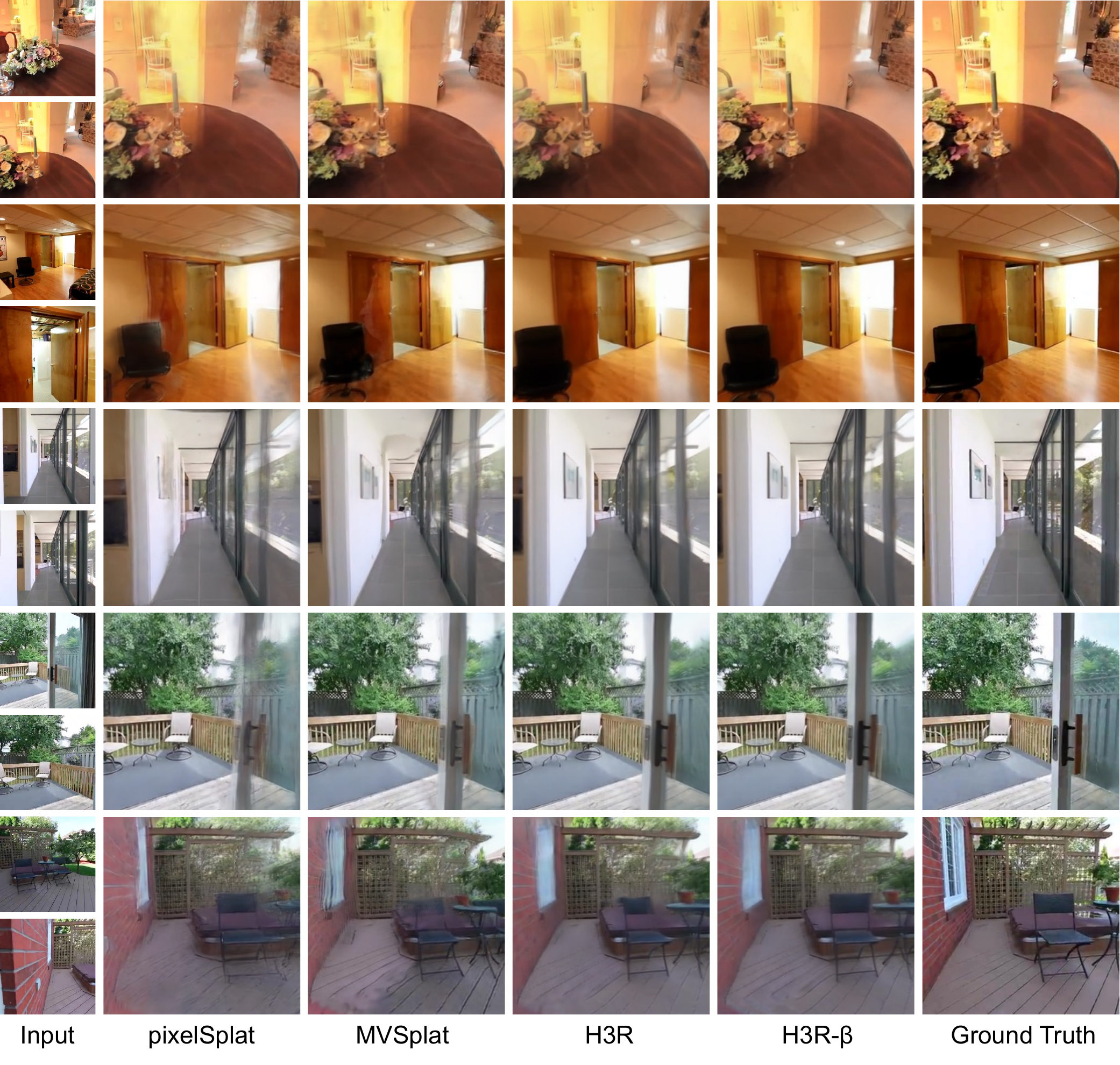}
    \caption{
        \textbf{Impact of input resolution on RealEstate10K.}
        Using 512$\times$512 inputs, our H3R-$\beta$ achieves more accurate geometry and photorealistic texture than recent methods, particularly for sharp edges and complex surfaces.
    }
    \label{fig:high_resolution}
\end{figure*}

We present additional qualitative results on RealEstate10K in~\cref{fig:target_pose,fig:view_scaling_visual,fig:high_resolution}.
Collectively, these studies illustrate that incorporating target pose, more input views, and higher resolution inputs directly contributes to substantial gains in structural integrity, detail preservation, and overall photorealism.

\begin{table*}[!htbp]
    \renewcommand{\arraystretch}{1.1} 
    \centering
    \begin{tabular}{l|ccc}
        \toprule
        \textbf{config}      & \textbf{H3R}                                                                 & \textbf{H3R-$\alpha$} & \textbf{H3R-$\beta$} \\
        \midrule
        peak learning rate   & 1e-4                                                                         & 5e-5                  & 5e-5                 \\
        min learning rate    & 5e-5                                                                         & -                     & -                    \\
        warm-up steps        & 3,000                                                                        & 0                     & 0                    \\
        LR schedule          & \begin{tabular}[c]{c}cosine decay\\ 150k steps,\\ then constant\end{tabular} & constant              & constant             \\
        \midrule
        optimizer            & \multicolumn{3}{c}{Adam}                                                                                                    \\
        betas                & \multicolumn{3}{c}{(0.9, 0.999)}                                                                                            \\
        weight decay         & \multicolumn{3}{c}{0}                                                                                                       \\
        gradient clip        & \multicolumn{3}{c}{0.5}                                                                                                     \\
        total batch size     & \multicolumn{3}{c}{16}                                                                                                      \\
        \midrule
        EMA decay            & \multicolumn{3}{c}{0.999}                                                                                                   \\
        \midrule
        trainable parameters & \multicolumn{3}{c}{90.9 M}                                                                                                  \\
        training steps       & 1,000,000                                                                    & 30,000                & 80,000               \\
        GPU                  & 4 $\times$ RTX 4090                                                          & 4 $\times$ A100-80GB  & 4 $\times$ A100-80GB \\
        training time        & 7.4 days                                                                     & 15 hours              & 42 hours             \\
        \bottomrule
    \end{tabular}
    \caption{
        \textbf{Training settings for RealEstate10K.}
    }
    \label{tab:training_setting_re10k}
\end{table*}
\begin{table*}[!htbp]
    \renewcommand{\arraystretch}{1.1} 
    \centering
    \begin{tabular}{l|ccc}
        \toprule
        \textbf{config}      & \textbf{H3R}                                                                 & \textbf{H3R-$\alpha$} & \textbf{H3R-$\beta$} \\
        \midrule
        peak learning rate   & 1e-4                                                                         & 5e-5                  & 5e-5                 \\
        min learning rate    & 5e-5                                                                         & -                     & -                    \\
        warm-up steps        & 3,000                                                                        & 0                     & 0                    \\
        LR schedule          & \begin{tabular}[c]{c}cosine decay\\ 150k steps,\\ then constant\end{tabular} & constant              & constant             \\
        \midrule
        optimizer            & \multicolumn{3}{c}{Adam}                                                                                                    \\
        betas                & \multicolumn{3}{c}{(0.9, 0.999)}                                                                                            \\
        weight decay         & \multicolumn{3}{c}{0}                                                                                                       \\
        gradient clip        & \multicolumn{3}{c}{0.5}                                                                                                     \\
        total batch size     & \multicolumn{3}{c}{16}                                                                                                      \\
        \midrule
        EMA decay            & \multicolumn{3}{c}{0.999}                                                                                                   \\
        \midrule
        trainable parameters & \multicolumn{3}{c}{90.9 M}                                                                                                  \\
        training steps       & 400,000                                                                      & 90,000                & 20,000               \\
        GPU                  & 4 $\times$ RTX 4090                                                          & 8 $\times$ RTX 4090   & 4 $\times$ A100-80GB \\
        training time        & 3 days                                                                       & 28 hours              & 11 hours             \\
        \bottomrule
    \end{tabular}
    \caption{
        \textbf{Training settings for ACID}.
    }
    \label{tab:training_setting_acid}
\end{table*}

\end{document}